\documentclass[conference]{IEEEtran}
\usepackage{tikz}
\usetikzlibrary{positioning}

\usepackage{booktabs}  % For better horizontal rules
\usepackage{siunitx}   % For better number formatting
\usepackage{xcolor}    % For adding color
\usepackage{tikz}
\usetikzlibrary{positioning, shadows, decorations.pathreplacing, backgrounds}
\usepackage{booktabs}
\usepackage{siunitx}
\usepackage{colortbl}
\usetikzlibrary{positioning, arrows.meta}

\usepackage{tikz}
\usetikzlibrary{patterns}
\usepackage{amsmath}
\usepackage[multiple]{footmisc}
\usepackage{booktabs}
\usepackage{amsthm}
\usepackage{dsfont}

\usepackage{amsmath}

\usepackage[utf8]{inputenc}
\usepackage{balance}
\usepackage{siunitx}
\usepackage{multirow}
\usepackage{graphicx}
\usepackage{listings}
\usepackage{commath}
\usepackage{epstopdf}
\usepackage{color}
\usepackage{url}
\usepackage{caption}

\usepackage{alltt}
\usepackage{bbm}
\usepackage{mathrsfs}
\usepackage{float}
\usepackage{subcaption}
\usepackage{graphicx,fancyvrb}
\usepackage[linesnumbered,ruled,vlined]{algorithm2e}
\usepackage{algpseudocode}

\usepackage{mathtools}
\usepackage{capt-of}
\usepackage{colortbl}
\usepackage{xcolor}
\usepackage{xfrac}
\usepackage{footnote}

\usepackage{tikz}
\usetikzlibrary{shapes, arrows, positioning}

 \usepackage{enumitem}
\usepackage{array}
\usepackage{arydshln}
\definecolor{mygreen}{rgb}{0,0.6,0}
\definecolor{myred}{rgb}{0.6,0,0}
\definecolor{mygray}{rgb}{0.5,0.5,0.5}
\definecolor{mymauve}{rgb}{0.58,0,0.82}
\definecolor{myblue}{rgb}{0,0,1}

\usepackage{listings}
\lstnewenvironment{VerbatimText}[1][]{
    
    \lstset{fancyvrb=true,basicstyle=\footnotesize,captionpos=b,xleftmargin=2em,#1}
}{}

\PassOptionsToPackage{hyphens}{url}\usepackage{hyperref}

\DeclareMathAlphabet\mathbfcal{OMS}{cmsy}{b}{n}

\definecolor{ReviewerOneColor}{RGB}{240, 110, 0}   % Orange
\definecolor{ReviewerTwoColor}{RGB}{0, 139, 109}   % Dark Cyan
\definecolor{ReviewerThreeColor}{RGB}{45, 85, 205}
\definecolor{MetaReviewerColor}{RGB}{142, 69, 133} % Plum

\definecolor{darkred}{rgb}{0.55, 0.0, 0.0}

\newcommand{\starter}[1]{\vspace{2mm}\noindent {\bf #1:}}

\newcommand{\EOD}{\text{EOD}\xspace}
\newcommand{\EO}{\text{EO}\xspace}
\newcommand{\DP}{\text{DP}\xspace}

\newcommand{\relation}{D\xspace}
\newcommand{\equivPairs}{M\xspace}
\newcommand{\allTuples}{\mathcal{D}\xspace}
\newcommand{\pairs}{P\xspace}
\newcommand{\tuple}{t}
\newcommand{\schema}{\mathcal{S}\xspace}
\newcommand{\attr}{A\xspace}

\newcommand{\RR}{\text{RR}\xspace}
\newcommand{\PQ}{\text{PQ}\xspace}
\newcommand{\PC}{\text{PC}\xspace}
\newcommand{\Fb}{\ensuremath{\text{F}_{\PC,\RR}}\xspace}

\newcommand{\walamz}{\texttt{WAL-AMZ}\xspace} % walmart-amazon
\newcommand{\amzgog}{\texttt{AMZ-GOO}\xspace} % amazon- google
\newcommand{\DBLPGogS}{\texttt{DBLP-GOO}\xspace} % dblp - google scholar
\newcommand{\DBLPACM}{\texttt{DBLP-ACM}\xspace} % dblp- acm

\newcommand{\itunamz}{\texttt{ITU-AMZ}\xspace} % itunes-amazon
\newcommand{\FodorZag}{\texttt{FOD-ZAG}\xspace} % fodors - zagat
\newcommand{\Beer}{\texttt{BEER}\xspace} % Beer
\newcommand{\candidatePairs}{C}

\newcommand{\dom}[1]{\textsc{Dom}(#1)}
\newcommand{\boxtheorem}{\hfill $\blacksquare$\vspace{2mm}}

\newcommand{\ignore}[1]{}

\definecolor{black}{rgb}{0,0,0}
\definecolor{grey}{rgb}{0.8,0.8,0.8}
\definecolor{red}{rgb}{1,0,0}
\definecolor{green}{rgb}{0,1,0}
\definecolor{darkgreen}{rgb}{0,0.5,0}
\definecolor{darkpurple}{rgb}{0.5,0,0.5}
\definecolor{darkdarkpurple}{rgb}{0.3,0,0.3}
\definecolor{blue}{rgb}{0,0,1}
\definecolor{shadegreen}{rgb}{0.95,1,0.95}
\definecolor{shadeblue}{rgb}{0.95,0.95,1}
\definecolor{shadered}{rgb}{1,0.85,0.85}
\definecolor{shadegrey}{rgb}{0.85,0.85,0.85}
\definecolor{oddRowGrey}{rgb}{0.80,0.80,0.80}
\definecolor{evenRowGrey}{rgb}{0.85,0.85,0.85}
\definecolor{lightpurple}{rgb}{0.88,1.0,1.0}

\newcommand{\RNum}[1]{\uppercase\expandafter{\romannumeral #1\relax}}
\newtheorem{defn}{Definition}[section]

\newtheorem{exa}[defn]{Example}
\newtheorem{col}[defn]{Colollary}

\newcommand{\proj}[1]{{\Pi}}
\newcommand{\sel}[1]{{\sigma}}

\newcommand{\cut}[1]{}
\newcommand{\eat}[1]{}

\usepackage{xcolor}

\definecolor{darkred}{rgb}{0.55, 0.0, 0.0}
\definecolor{darkgreen}{rgb}{0.0, 0.5, 0.0}

\usepackage[multiple]{footmisc}

\SetKwInput{KwInput}{Input}
\SetKwInput{KwOutput}{Output}

\usepackage{siunitx}
\ExplSyntaxOn
  \cs_new_eq:NN \calc \fp_eval:n
\ExplSyntaxOff

\definecolor{cbgreen}{RGB}{0,158,115} % Bluish-green
\definecolor{cbred}{RGB}{213,94,0} % Vermillion, a reddish-orange that is often recommended for being colorblind-friendly

\definecolor{DodgerBlue}{RGB}{30, 144, 255}

\usepackage{amsthm}

\usepackage{enumitem}

\newcommand{\AutoBlock}{\texttt{AUTO}\xspace}
\newcommand{\CTT}{\texttt{CTT}\xspace}
\newcommand{\graphblk}{\texttt{GRAPH}\xspace}
\newcommand{\suffix}{\texttt{Suffix}\xspace}
\newcommand{\exSuffix}{\texttt{XSuffix}\xspace}
\newcommand{\qgram}{\texttt{QGram}\xspace}
\newcommand{\exQgram}{\texttt{XQGram}\xspace}
\newcommand{\stdBlock}{\texttt{StdBlck}\xspace}

\IEEEoverridecommandlockouts

\SetKwInput{KwInput}{Input}
\SetKwInput{KwOutput}{Output}

\def\BibTeX{{\rm B\kern-.05em{\sc i\kern-.025em b}\kern-.08em
    T\kern-.1667em\lower.7ex\hbox{E}\kern-.125emX}}

\begin{document}

\title{Evaluating Blocking Biases in Entity Matching}

\author{

\IEEEauthorblockN{Mohmmad Hossein Moslemi}
\IEEEauthorblockA{\textit{The University of Western Ontario} \\
\textit{London, Ontario, Canada}\\
mohammad.moslemi@uwo.ca}
\and

\IEEEauthorblockN{Harini Balamurugan}
\IEEEauthorblockA{\textit{The University of Western Ontario} \\
\textit{London, Ontario, Canada}\\
hbalamur@uwo.ca}
\and

\IEEEauthorblockN{Mostafa Milani}\IEEEauthorblockA{\textit{The University of Western Ontario} \\
\textit{London, Ontario, Canada}\\
mostafa.milani@uwo.ca}
}

\maketitle

\begin{abstract}
Entity Matching (EM) is crucial for identifying equivalent data entities across different sources, a task that becomes increasingly challenging with the growth and heterogeneity of data. Blocking techniques, which reduce the computational complexity of EM, play a vital role in making this process scalable. Despite advancements in blocking methods, the issue of fairness—where blocking may inadvertently favor certain demographic groups—has been largely overlooked. This study extends traditional blocking metrics to incorporate fairness, providing a framework for assessing bias in blocking techniques. Through experimental analysis, we evaluate the effectiveness and fairness of various blocking methods, offering insights into their potential biases. Our findings highlight the importance of considering fairness in EM, particularly in the blocking phase, to ensure equitable outcomes in data integration tasks.
\end{abstract}
\section{Introduction}\label{sec:intro}

Entity Matching (EM) involves determining whether two or more data entities from the same or different sources refer to the same real-world object. Also known as entity linkage or record matching, EM is crucial for data integration and has extensive applications across various industries ~\cite{mudgal2018deep, fu2021hierarchical, li2020deep, yao2022entity, konda2018magellan\ignore{, simonini2018schema}}. EM methods typically consist of a matching component that compares entities and labels them as either ``match'' or ``non-match.'' A major challenge in EM is its computational complexity, often scaling quadratically as each entity must be compared to all others, making it an $O(n^2)$ problem. This complexity becomes prohibitive for large datasets with millions of records. To mitigate this, blocking methods are used as a preliminary step to reduce the number of comparisons~\cite{michelson2006learning, \ignore{bilenko2006adaptive, }ebraheem2018distributed}. As a result, EM systems typically operate in two phases: blocking to limit comparisons, followed by matching to produce the final labels.

Blocking reduces the number of entity comparisons by grouping similar entities into distinct or overlapping blocks, thereby limiting comparisons to smaller, more manageable groups. This approach is essential for improving the scalability of EM by addressing its quadratic complexity and reducing computational load. Over time, blocking methods have evolved from simple heuristic-based techniques to more advanced methods. Traditional methods, such as Standard Blocking and Sorted Neighborhoods~\cite{papadakis2020blocking, li2020survey}, laid the groundwork for grouping entities. Recent advancements include techniques that utilize machine learning and deep learning for greater efficiency, such as Canopy Clustering~\cite{mccallum2000efficient}, and modern frameworks that optimize attribute selection and candidate generation~\cite{michelson2006learning, bgp, zhang2020autoblock, ebraheem2018distributed}.

In recent years, fairness in ML has gained significant attention ~\cite{zafar2017fairness, hardt2016equality, dwork2012fairness} due to its critical impact on real-life applications. Fairness is particularly important in the context of EM because both EM and blocking systems can produce biased results, often exhibiting higher accuracy for one demographic group over another. Despite the significant implications of EM on real-life decisions, research on the fairness of EM remains limited, with only a few studies exploring this issue ~\cite{efthymiou2021fairer,nilforoushan2022entity,shahbazi2023through, moslemi2024threshold}. Even fewer studies have addressed the fairness of blocking methods, leaving this area largely unexplored. To the best of our knowledge, only one study has investigated the fairness of blocking~\cite{shahbazi2024fairness}, which merely touched on the topic by defining a fairness metric similar to the Reduction Ratio disparity, one of the metrics we consider, and proposing simple algorithms to address bias based on that measure. However, focusing solely on this metric is insufficient, as it overlooks other critical aspects of fairness. This highlights a significant gap in the literature, as fairness issues in blocking could lead to disproportionate group representation or exclusion, amplifying biases in downstream matching tasks.

In this paper, we investigate fairness in blocking methods for EM. Traditional fairness metrics, such as Equalized Odds, Equal Opportunity, and Demographic Parity~\cite{hardt2016equality, dwork2012fairness}, are typically used for classification tasks in machine learning models and do not apply to blocking methods. We extend existing blocking metrics, including Reduction Ratio (\RR), Pair Completeness (\PC), and their harmonic mean (\Fb), to define biases in blocking methods. Our study evaluates these extended metrics to assess bias across various blocking techniques. We conduct an experimental analysis to identify fairness issues, detect biases, and understand their impact on EM and end-to-end matching tasks. The aim is to uncover biases, understand their origins, and develop debiasing methods that ensure blocking methods are both accurate and fair.

The paper is structured as follows. Section \ref{sec:rw} presents related work, including a summary of the state-of-the-art blocking methods. Section \ref{sec:Background} provides a formal definition of EM and blocking and reviews the metrics used to evaluate blocking. In Section \ref{sec:blocking-bias}, we extend these blocking metrics to fairness measures, proposing them as a sufficient method for assessing bias in blocking. Section \ref{sec:exp} presents our experimental results, highlighting the effectiveness of various blocking methods and the biases they may introduce. Section \ref{sec:conclusion} offers our conclusions. All the implementations are available at \url{https://github.com/mhmoslemi2338/pre-EM-bias}.

\section{Related Work} \label{sec:rw}

We briefly review the existing blocking methods for EM and then discuss fairness in EM. 

\subsection{Blocking Methods}\label{sec:blocking-rw}

Over the years, a wide range of blocking techniques has been developed, from simple heuristic-based methods to advanced approaches involving deep learning (for surveys, see ~\cite{papadakis2016comparative, christen2011survey, papadakis2020blocking}). These methods can be categorized in various ways, each providing a different perspective on their function and applications. One way is to distinguish between learning-based and non-learning-based algorithms. Rule-based methods, a type of non-learning-based approaches, rely on expert knowledge or simple heuristics to define the blocking criteria. In contrast, learning-based methods require training data to learn how to block the data using machine learning techniques.

Another categorization is based on schema awareness. Schema-aware methods focus on the most important attributes of the data, while schema-agnostic methods treat the entire entity as a single attribute, utilizing all available information. A third categorization concerns redundancy awareness. It divides methods into three subcategories: redundancy-free, redundancy-positive, and redundancy-neutral. These subcategories differ in how they handle the assignment of entities to blocks and the overlap between blocks~\cite{papadakis2016comparative}. Redundancy-free methods create disjoint blocks by assigning each entity to only one block. In these methods, similarity between two entities increases if they share a greater number of identical blocks. In contrast, redundancy-positive methods form overlapping blocks, with each entity placed in multiple blocks; here, the likelihood of a match between two entities is proportional to the total number of blocks they share. Redundancy-neutral methods also produce overlapping blocks, but in such a way that most entity pairs share nearly the same number of blocks. As a result, in these methods, redundancy is insignificant for determining similarity between entities. These categorizations illustrate the range of blocking techniques. Each is tailored to specific scenarios based on data characteristics and the trade-off between computational efficiency and matching accuracy.

Traditional blocking methods such as Standard Blocking~\cite{fellegi1969theory} and Sorted Neighborhood~\cite{hernandez1995sorted} have fundamentally shaped the field. Standard Blocking categorizes records according to a blocking key, such as a phone number or surname initials, to conduct intensive comparisons within these blocks. However, this method risks inefficiencies when block sizes are large. In contrast, Sorted Neighborhood enhances efficiency by sorting records according to a key and employing a sliding window for comparisons, though it may overlook matches when key values exceed the window's boundaries.

Advanced approaches like Canopy Clustering~\cite{mccallum2000efficient} have been developed to overcome some limitations of traditional methods. This technique uses a low-cost, coarse similarity measure to group records, followed by more precise and computationally intensive comparisons within each canopy. While it aims to reduce the total number of comparisons, it can sometimes incorrectly group distinct entities together~\cite{bgp}.

Blocking techniques, such as BSL (Blocking Scheme Learner)~\cite{michelson2006learning} and BGP (Blocking based on Genetic Programming)~\cite{bgp}, have further enhanced the efficiency and accuracy of the traditional methods. These more advanced methods leverage machine learning to refine blocking schemes, focusing on attribute selection and comparison methods to generate blocks efficiently. CBLOCK offers an automated approach to canopy formation within a map-reduce framework, tailored specifically for large-scale de-duplication tasks involving diverse datasets. It optimizes the trade-off between recall and computational efficiency~\cite{sarma2011cblock}. 

Deep learning has recently revolutionized the blocking phase in EM, shifting from traditional heuristic methods to more adaptive and automated approaches. Frameworks like AutoBlock and DeepER exploit deep learning for representation learning and nearest neighbor search, showing significant effectiveness across large-scale datasets ~\cite{zhang2020autoblock, ebraheem2018distributed}. DeepBlock, which merges syntactic and semantic similarities, further enhances blocking quality by accurately grouping similar records, even in noisy or heterogeneous datasets ~\cite{javdani2019deepblock}.

Various metrics have been used to evaluate blocking quality, but three are most commonly considered comprehensive: \RR, \PQ, and \PC~\cite{christen2007quality, elfeky2002tailor, christen2011survey}. \RR measures how much a blocking method reduces the total number of comparisons, \PQ represents the percentage of candidate pairs that are true matches after blocking, and \PC indicates the percentage of true matches present in the candidate set after blocking.

There is little agreement in the literature on the evaluation measures for blocking. Some studies only considered \PC and \RR as their primary metrics~\cite{zhang2020autoblock, thirumuruganathan2021deep, ebraheem2018distributed, papadakis2011efficient, michelson2006learning,\ignore{bilenko2006adaptive, }steorts2014comparison}, while some only considered \PC and \PQ~\cite{javdani2019deepblock, de2009robust}. Several studies also considered runtime as a main evaluation measure in blocking~\cite{galhotra2021efficient, zeakis2023pre, sarma2011cblock}. There is also research that employed the harmonic mean of \PC and \RR~\cite{o2019review, o2018new, kopcke2010frameworks, kejriwal2013unsupervised, mugeni2023graph} or the harmonic mean of \PC and \PQ~\cite{christen2011survey}. We will formally define these metrics in Section \ref{sec:Background} and discuss which measures provide a more meaningful way to evaluate the blocking methods considered in Section~\ref{sec:exp}.

\subsection{Fairness in Entity Matching} \label{sec:fair-em-rw}

Fairness is a growing concern in EM systems, especially as these systems are widely employed in data-driven decision-making processes. A key concern is that EM processes can unintentionally perpetuate biases present in the data, leading to unfair outcomes. A real-world example that highlights these fairness issues is the use of ``banned lists.'' Individuals, particularly people of color, have been wrongly matched with entries on the list and flagged as security threats due to misidentification. This has led to the denial of services~\cite{acluNoFly}. Another example is Google's advertising algorithms, which have been found to match and display prestigious job advertisements more frequently to men than to women. This bias occurs because algorithms tend to match male profiles more closely with the job requirements~\cite{googleJobAdsBias}. These discrepancies highlight how EM algorithms can perpetuate existing gender biases, resulting in unequal opportunities and outcomes. These examples illustrate the necessity for developing end-to-end matching algorithms (including blocking) that ensure fair and equitable treatment across all demographic groups.

Recent research has aimed to address fairness challenges in matching methods. The FairER algorithm, introduced by~\cite{efthymiou2021fairer}, incorporates fairness constraints directly into EM methods, emphasizing the importance of considering fairness from the outset of the data cleaning pipeline. Additionally,~\cite{nilforoushan2022entity} proposed an AUC-based fairness metric to evaluate the performance of EM methods that return matching scores, assessing their effectiveness across different groups. Recent studies by~\cite{shahbazi2024fairness} and~\cite{shahbazi2023through} have explored biases in EM as a data preparation task, aiming to prevent the disproportionate exclusion or misrepresentation of certain groups. However, these works primarily focus on EM and offer limited discussion on the fairness of blocking methods.

\section{Background} \label{sec:Background}

We start from a relation schema $\schema$ consisting of a set of attributes $\attr_1,...,\attr_m$ with domains $\dom{\attr_i}, i\in [1,m]$. An entity (record) $\tuple$ with schema $\schema$ is a member of $\dom{\attr_1}\times...\times \dom{\attr_m}$, the set of all possible entities, which we denote by $\allTuples$. We use $\tuple[\attr_i]$ to refer to the value of the attribute $\attr_i$ in an entity $\tuple$. A relation $\relation \subseteq \allTuples$ is a set of entities. 

EM generally has two major phases: blocking and matching~\cite{mudgal2018deep}. We begin with matching and then explain blocking.

\subsection{Entity Matching} \label{sec:entity-matching}
Given two relations $D_1,D_2$, the problem of \emph{entity matching (EM)} is to find a subset $\equivPairs$ of $D_1 \times D_2$ that consists of entity pairs referring to the same real-world entities. We refer to such entity pairs as {\em equivalent pairs}. An entity matcher (matcher in short) for entities of schema $\schema$ is a binary classifier $f: \allTuples\times \allTuples \mapsto \{0,1\}$; which labels entity pairs $1$, meaning they match, or $0$, meaning they don't match. Given relations $D_1,D_2$ with schema $\schema$ and equivalent entity pairs $\equivPairs \subseteq D_1\times D_2$, the goal of $f$ is to find the equivalent entities in $\equivPairs$, i.e., label the entity pairs in $\equivPairs$ as a match and the non-equivalent entities as non-match. The accuracy of $f$, e.g., accuracy rates and F1 score, is defined based on how correctly it labels the equivalent pairs.

%A matcher $f$ is often derived from a score function $s: \allTuples\times \allTuples \mapsto [0,1]$ and a classification threshold $\threshold \in [0,1]$, where $f(\rPair)=\mathbbm{1}_{s(\rPair) \ge \threshold}$ for every pair $\rPair\in \allTuples\times \allTuples$ ($\mathbbm{1}$ is the indicator function). For a score function $s$ and a matching threshold $\threshold$, we define accuracy using a probability distribution $\pr$ of random entity pair $\st{X} \in \allTuples\times \allTuples$; {\em true-positive rate (\TPR)}, $\TPR(s, \threshold)$, is $\pr(s(\st{X}) \ge \threshold \mid \st{X} \in \equivPairs)$, and \TNR, \FPR, \FNR, accuracy, and F1 measures are defined similarly using $\pr$. The AUC for the matching score $s$ is defined as $\AUC(s)=\int_{0}^{1} \TPR(s, \FPR^{-1}(s, \threshold))\;d \threshold$ for varying threshold $\threshold \in [0,1]$. It quantifies the likelihood that an equivalent pair is accurately ranked above a non-equivalent pair, offering a measure of matching accuracy that does not depend on any specific matching threshold $\threshold$. It is estimated using the trapezoidal rule~\cite{fawcett2006introduction} or Wilcoxon rank-sum test~\cite{hanley1982meaning}.

\subsection{Blocking} The set of entity pairs for relations $\relation_1,\relation_2$ in a matching setting, which we denote by $\pairs=\relation_1 \times \relation_2$, grows quadratically in size w.r.t the size of $\relation_1$ and $\relation_2$. This makes it costly to run expensive matching methods for all possible pairs. The problem of blocking is to find a candidate set $\candidatePairs \subseteq \pairs$, which is much smaller than $\pairs$, while it still includes all equivalent pairs; $\equivPairs \subseteq \candidatePairs$. A blocking method is expected to run much faster than a matching method for comparing all possible pairs. Blocking saves unnecessary checking of some of the non-equivalent pairs while searching for the equivalent pairs. 

\begin{defn}[Blocking]
Given two datasets, $\relation_1$ and $\relation_2$ with the set of all possible pairs $\pairs = \relation_1 \times \relation_2$, the goal of blocking is the fast computation of a candidate set $\candidatePairs$ s.t. $|\candidatePairs| \ll |\pairs|$ (there are much fewer candidates compared to total pairs) and $\equivPairs \subseteq \candidatePairs$.
\end{defn}

\starter{Blocking Metrics} How effective a blocking method is measured using three main quality measures defined as follows: \begin{align}
    \RR=1 - \frac{|\candidatePairs|}{|\pairs|}\hspace{0.5cm}
    \PC=\frac{|\candidatePairs \cap \equivPairs|}{|\equivPairs|}\hspace{0.5cm}\PQ = \frac{|\candidatePairs\cap \equivPairs|}{|\candidatePairs|}
\end{align}

\noindent \emph{Reduction ratio} (\RR) is the ratio of the reduction in the number of comparisons after blocking to the total number of possible comparisons. A higher \RR value signifies a greater reduction in the number of candidate entity pairs. This measure does not consider the quality of the generated candidate entity pairs. \emph{Pair completeness} (\PC) represents the ratio of equivalent pairs retained after blocking to the total number of equivalent pairs. This measure evaluates how effectively $\candidatePairs$ preserves equivalences, corresponding to recall in information retrieval~\cite{hernandez1998real}. \PC ranges from 0 to 1, where a value of 1 indicates that the blocker retains all true matches. {\em Pairs quality} (\PQ) denotes the fraction of equivalent pairs produced by $\candidatePairs$ relative to the total number of pairs. A higher \PQ indicates that $\candidatePairs$ is efficient, primarily generating true matches. In contrast, a lower \PQ suggests that many non-matching pairs are included. \PQ is equivalent to precision in information retrieval~\cite{papadakis2020blocking}. \PQ ranges from 0 to 1, with higher values indicating that the blocker is more effective at eliminating non-matches.

Among these metrics, \PQ is typically sensitive to small changes in \RR. In large datasets with millions of pairs and a much smaller number of equivalent pairs, even a small change in \RR can significantly increase the size of the candidate set. This, in turn, greatly impacts \PQ, making it highly sensitive. Therefore, in most studies, \PC, \RR, and their harmonic mean are used as the key metrics for evaluating blocking methods~\cite{papadakis2016comparative, zeakis2023pre, papadakis2011efficient, kejriwal2013unsupervised}. The harmonic mean, which balances the trade-off between \PC and \RR, is defined as: \begin{align}
\Fb = \frac{2 \times \PC \times \RR}{\PC + \RR}
\end{align}

In addition to evaluating the effectiveness of blocking methods, their efficiency is equally important. Specifically, the runtime of a blocking method plays a crucial role in determining the optimal blocking method. A detailed numerical analysis of these measures and their relationships is provided in Section~\ref{sec:exp}. We explain these measures using an example.

\begin{exa} \label{ex:blocking} Figure~\ref{fig:blocking} shows a relation $\relation$ with ten entities. We consider matching entities in the same relation with entity pairs $\pairs=\relation\times\relation$. The red dotted lines show the equivalent pairs; $M=\{(t_2,t_6)$, $(t_6,t_7)$, $(t_2,t_7)$, $(t_4,t_5)$, $(t_{8},t_{9})$, $ (t_{8},t_{10})$, $(t_{9},t_{10})\}$. The solid black lines show a blocking method that specifies three blocks that result in $3+6+3=12$ candidate pairs. The blocking result misses two equivalent pairs, $(t_2, t_6)$ and $(t_2, t_7)$, but reduces the number of pairs to check from 45 to 12, resulting in \(\RR = 1 - \frac{12}{45} \approx 0.73\), \(\PC = \frac{5}{7} \approx 0.71\), and \Fb$ \ignore{=\frac{2\times\frac{5}{7}\times\frac{33}{45}}{\frac{5}{6} + \frac{33}{45} } = \frac{55}{76} }\approx 0.72$. 
% Additionally, \(\RR_{opt} = 1 - \frac{6}{45} \approx 0.87\). Using these values, $\NRRD = 1 - \sqrt{\frac{0.14}{0.87}} \approx 0.60$ and $\PBS = \sqrt{0.42 \times 0.60} \approx 0.50.$
\end{exa}

%\starter{Matching} The primary objective of the matching component in the EM system is to identify pairs and produce a label or score that reflects the likelihood of the pair being ``equivalent.'' The aim is to achieve optimal matching accuracy. However, this study does not focus on the matching aspect ~\cite{mudgal2018deep}.

% \begin{definition}[Matching]
% Given a arbitrary pair denoted by $p = (a,b)$ s.t. $a \in \relation_1$ and $b \in \relation_2$ the objective of a marching system is to gice score for each p in a way that maximizes the accuracy measure that is chosen based on the applicaton.
% \end{definition}

%\begin{defn}[Matching] Given an arbitrary pair \( p = (a,b) \), where \( a \in \relation_1 \) and \( b \in \relation_2 \), the goal of a matching system is to assign a score to each pair \( p \) in a manner that maximizes a chosen accuracy measure, tailored to the specific application. \end{defn}

% \subsection{Fairness} 
% \section{Preliminaries and Problem Definition}\label{sec:problem}

% Blocking methods can suffer from disparities; the quality of blocking might differ for the minority group compared with the majority, leading either to missing equivalent minority pairs or being ineffective in reducing unnecessary matching of non-equivalent methods. 

\section{Measuring Bias in Blocking} \label{sec:blocking-bias}

Blocking methods can suffer from disparities, with blocking quality differing for minority group compared to the majority, potentially missing equivalent minority pairs or failing to reduce unnecessary matching of non-equivalent pairs.

To define biases in blocking, we assume the relation schema $\schema$ includes a sensitive (or protected) attribute $\attr$ (e.g., gender or ethnicity) with domains $\dom{\attr}={a,b}$, where $a$ and $b$ correspond to the minority (e.g., female) and majority groups (e.g., male), respectively. This means that an entity $t$ with $t[\attr]=a$ belongs to the minority group, while $t[\attr]=b$ indicates it belongs to the majority group. In the context of matching, we use the protected attribute of entities in a record pair to determine whether the pair concerns minority or majority groups. We define a minority pair as any pair of entities $(t_1,t_2) \in \relation_1 \times \relation_2$ where at least one entity belongs to the minority group based on the specified sensitive attribute (e.g., ethnicity, gender). Conversely, a majority pair consists of both entities from the majority group. Intuitively, minority pairs could result in an EM decision that negatively affects a minority entity. This can occur if the entity is incorrectly matched with another minority or majority entity or if the opportunity to match it correctly with another entity is missed.

To define blocking disparities, we use $\pairs_g$, $\candidatePairs_g$, and $\equivPairs_g$ with $g\in \{a,b\}$ to respectively refer to the set of pairs in group $g$, and the set of candidate pairs in group $g$, and the set of equivalent pairs in group $g$. Then we define the reduction ratio and pair completeness for the group $g$ as follows:\begin{align}
    \RR_g=1-\frac{|\candidatePairs_g|}{|\pairs_g|} \hspace{1cm} \PC_g=\frac{|\candidatePairs_g \cap \equivPairs|}{|\equivPairs_g|}
\end{align}

\noindent Intuitively, the reduction ratio per group $g$ specifies the reduction in the number of comparisons for the entity pairs in group $g$. Similarly, pair completeness per group $g$ measures the completeness of comparisons within the pairs of group $g$ only. We do not define or use \PQ per group $g$, as explained earlier, due to its sensitivity and redundancy with \RR and \PC. Instead, using \RR and \PC per group, we define their harmonic mean for each group $g$ as $\Fb^g$.

\begin{figure}
    \centering
    \includegraphics[width=0.6\linewidth]{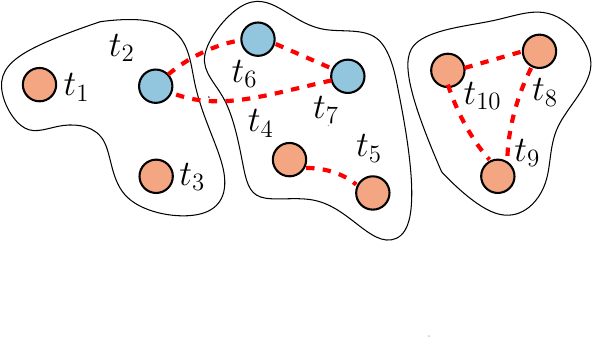}
    \vspace{-7mm}
    \caption{Disparity in blocking: minority and majority entities are highlighted in blue and red respectively, and the equivalent pairs are linked by dotted lines. Solid lines show the blocks.}
    \label{fig:blocking}
\end{figure}

% We define disparities as the differences between the blocking quality metrics for the majority and minority groups. For example, $\Delta\RR=\RR_b-\RR_a$ represents the reduction ratio disparity, $\Delta\PC=\PC_b-\PC_a$ denotes the pair completeness disparity, and $\Delta\Fb =\Fb^b-\Fb^a$ indicates the mean disparity. 

We define disparities as the differences between the blocking quality metrics for the majority and minority groups. For example, $\Delta\RR=\RR_b-\RR_a$ represents the \RR disparity, $\Delta\PC=\PC_b-\PC_a$ denotes the \PC disparity, and $\Delta\Fb =\Fb^b-\Fb^a$ indicates the \Fb disparity.

We now use our running example to explain these quality metrics per group $g$.
 
\begin{exa} \label{ex:fair-blocking} Continuing with the example in Figure~\ref{fig:blocking}, three entities belong to the minority group and seven to the majority group, as indicated by different colors. This results in $45 - 21 = 24$ minority pairs before blocking, where 45 is the total number of initial pairs. Using the combination formula, $\frac{7\times6}{2} = 21$ represents the total number of majority pairs (both entities belong to the majority group) before blocking. After blocking, we have five majority pairs and seven minority pairs (where at least one entity is from the minority group). The following are the blocking quality metrics per group: $\RR_a=1-7/24\approx 0.71$, $\PC_a=1/3\approx 0.33$, $\Fb^a\approx0.45$, $\RR_b=1-5/21\approx0.76$, $\PC_b=4/4=1$, and $\Fb^b\approx0.86$. These metrics give \RR, \PC, and \Fb disparities of $\Delta \RR\approx0.05$, $\Delta \PC\approx0.67$, and $\Delta \Fb\approx0.41$. In this example, all disparities are positive, indicating a lower blocking quality for the minority group $a$. This is evident in the missing equivalences for the minority pairs (2 missing pairs for the minority vs. none for the majority) and the smaller \RR for the minority group.
\end{exa}

Introducing these disparity metrics allows for evaluating the fairness of blocking techniques. The metric $\Delta \RR$ measures the difference in \RR between pairs associated with minority and majority groups. A lower \RR for a group can result in ineffective blocking for that group, leading to an increased number of non-equivalent pairs being passed to the matching process. This can cause the matching method to either fail to identify true matches or mistakenly match non-equivalent pairs. $\Delta \PC$ measures the difference in \PC and reveals biases in blocking quality. It indicates that more equivalent pairs are incorrectly excluded for a group, reducing the effectiveness of the overall matching process. $\Delta \Fb$ captures both types of biases by combining $\Delta \RR$ and $\Delta \PC$, thereby reflecting overall disparities in the blocking process. By quantifying these differences across demographic groups, we can identify potential biases and inequalities in the blocking procedure.

\section{Evaluation and Analysis} \label{sec:exp}

The purpose of our experiments is twofold: first, to assess the quality of existing blocking methods on the benchmarks used in the literature by comparing \RR, \PC, \Fb, and time (Sections~\ref{sec:runtime} and~\ref{sec:exp-blocking}). Second, we analyze the same methods for potential biases using the introduced disparity measures (Section~\ref{sec:exp-bias}, ~\ref{sec:bias-impact}, and~\ref{sec:exp-nosens}).

\subsection{Experimental Setup}

We briefly explain the datasets and the blocking methods used in this paper before presenting the experimental results.

\subsubsection{Datasets} 
We use datasets from prominent EM benchmarks: Amazon-Google (\amzgog), Walmart-Amazon (\walamz), DBLP-GoogleScholar (\DBLPGogS), DBLP-ACM (\DBLPACM), Beer (\Beer), Fodors-Zagat (\FodorZag), and iTunes-Amazon (\itunamz), as referenced in~\cite{mudgal2018deep,yang2020fairness}. The research community commonly adopts these datasets to evaluate EM system performance.

Consistent with prior studies ~\cite{nilforoushan2022entity,shahbazi2023through,moslemi2024threshold}, we categorize entities into minority and majority groups across various datasets. For instance, in \DBLPACM, including a female name in the ``authors'' attribute defines the minority. In \FodorZag, entities with the ``Type'' attribute exactly equal to ``Asian'' are classified as minority. In \amzgog, the presence of ``Microsoft'' in the ``manufacturer'' attribute defines a minority group. For \walamz, entities classified as ``printers'' under the ``category'' attribute are considered a minority group. In \DBLPGogS, entities are considered minority if the ``venue'' attribute includes ``vldb j.''.  For \Beer, entities with ``Beer Name'' containing the phrase ``red'' are classified as minority. Finally, in \itunamz, the minority group includes those where the ``Genre'' attribute contains the word ``Dance''. 

Detailed statistical information about these datasets is provided in Table~\ref{tab:datasets}. The numbers in parentheses refer to the corresponding parameter for the minority group. For example, $2.6k\;(96)$ for $|D_1|$ in \walamz means there are 2.6$k$ entities in $D_1$, with only $96$ of them belonging to the minority group. The majority group parameters can be inferred from the table.

\begin{table}[h!]
    \centering
    \Large
    \resizebox{0.5\textwidth}{!}{
        \begin{tabular}{p{1.9cm} >{\centering\arraybackslash}p{1cm} c c c c} % Narrow and centered second column
    \toprule
    \textbf{Dataset} &
    \#Attr. & % New colum
    \textbf{$|\relation_1|$} & 
    \textbf{$|\relation_2|$} &
    $|\pairs|$ &
    $|\equivPairs|$ \\
    \midrule
        \walamz &5& 2.6k (96) & 22.0k (172) & 56.4m (2.5m) & 962 (88) \\
        \Beer &4 & 4.3k (1.3k) & 3.0k (932) & 13.0m (6.8m) & 68 (29)  \\
        \amzgog &3 & 1.4k (83) & 3.2k (4) & 4.4m (272.9k) & 1.2k (60) \\
        \FodorZag & 6& 533 (72) & 331 (3) & 176.4k (25.2k) & 111 (10) \\
        \itunamz & 8& 6.9k (1.9k) & 55.9k (12.7k) & 386.2m (171.0m) & 132 (40)  \\
        \DBLPGogS &4 & 2.6k (191)& 64.3k (389) & 168.1m (13.2m) & 5.3k (403)  \\
        \DBLPACM & 4& 2.6k (251)& 2.3k (225)& 6.0m (1.1m) & 2.2k (310)  \\
    \bottomrule
    \end{tabular}
}
    \caption{Datasets and their characteristics.}
    \vspace{-2mm}
    \label{tab:datasets}
\end{table}

\subsubsection{Blocking Methods} The blocking methods used in our study are as follows:

\begin{enumerate}[left=0pt]
    \item \textit{Standard Blocking (\stdBlock)}: This hash-based method groups records using concatenated attribute values (blocking keys) to form redundancy-free blocks. However, it is sensitive to noise; any variation in the key may exclude matching records from the same block~\cite{fellegi1969theory}.

    \item \textit{Q-Grams (\qgram) and Extended Q-Grams (\exQgram)}: Q-grams blocking splits blocking keys into subsequences of \textit{q} characters, improving noise tolerance but potentially increasing block size and number. Extended Q-Grams further combines q-grams for more distinctive keys, reducing block size and enhancing efficiency~\cite{christen2011survey, papadakis2015schema}.

    \item \textit{Suffix (\suffix) and Extended Suffix (\exSuffix) Arrays Blocking}: Suffix Arrays Blocking converts blocking keys into suffixes of a minimum length to form blocks, filtering out common suffixes to prevent oversized blocks. Extended Suffix Arrays consider all substrings longer than the minimum length, boosting noise tolerance~\cite{christen2011survey, papadakis2015schema}.

    \item \textit{AutoEncoder (\AutoBlock) and Cross-Tuple Training (\CTT) Blocking}: These deep learning methods group similar records into blocks using embeddings. \AutoBlock generates embeddings via an autoencoder to handle diverse and noisy data, while \CTT uses a Siamese Summarizer to create embeddings from synthetic data, enhancing distinction between matching and non-matching tuples~\cite{mudgal2018deep}.

    \item \textit{Semantic-Based Graph Blocking (\graphblk)}: The algorithm concatenates entity attributes into a single string, from which a transformer-based model extracts context-aware embeddings. These embeddings are reduced to a lower-dimensional space for efficiency. A k-nearest neighbor algorithm then constructs a graph of database tuples, connecting nodes based on embedding proximity. Unsupervised clustering is applied to group similar entities, framing the blocking task as a graph clustering problem~\cite{mugeni2023graph}.    
\end{enumerate}

We selected these methods as they represent the best-performing approaches in their respective categories. \stdBlock, \qgram, and \suffix (along with their extended versions) are widely recognized traditional methods and remain commonly used in practice due to their effectiveness and efficiency. With the growing trend of employing deep learning in blocking, we included \AutoBlock and \CTT, which have demonstrated superior performance in handling complex and noisy data. We chose \graphblk, an innovative and recent approach, to represent cutting-edge advancements in the blocking problem.

\subsection{Experimental Results} 

Our experimental results include an analysis of blocking methods based on their performance—specifically, blocking quality and runtime—which aligns with our first experimental objective. Additionally, we extensively analyze the existing biases in these methods to address our second objective.

\subsubsection{Runtime and Scalability} \label{sec:runtime}
\begin{figure}[htb]
   \centering
   \subfloat[All datasets]{\label{fig:all-time}
			{\raisebox{+.04\height}{\includegraphics[width=0.22\textwidth]{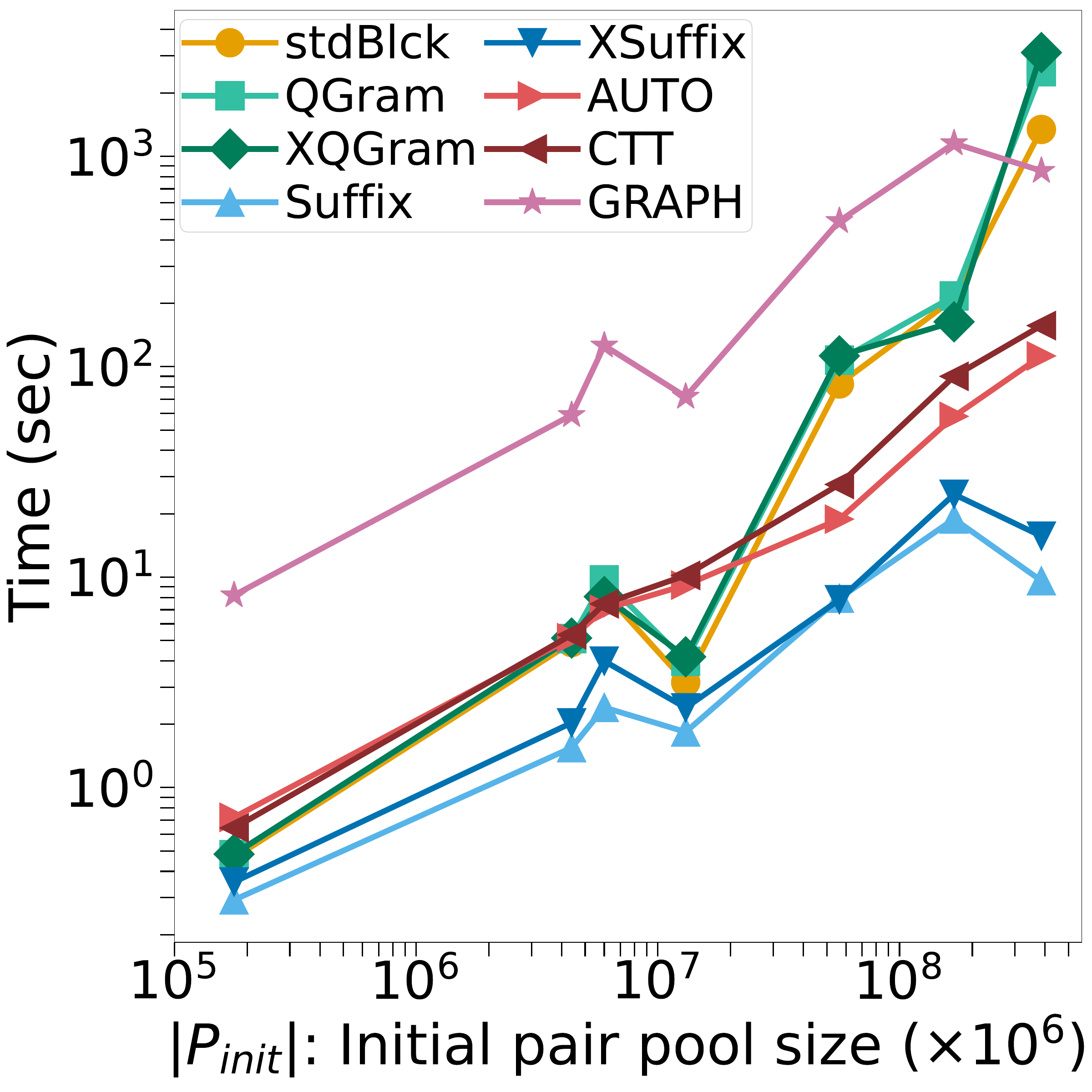}}}}	
   \subfloat[\walamz]{\label{fig:time_walamz}   
    {\raisebox{-.5\height}{\includegraphics[width=0.24\textwidth]{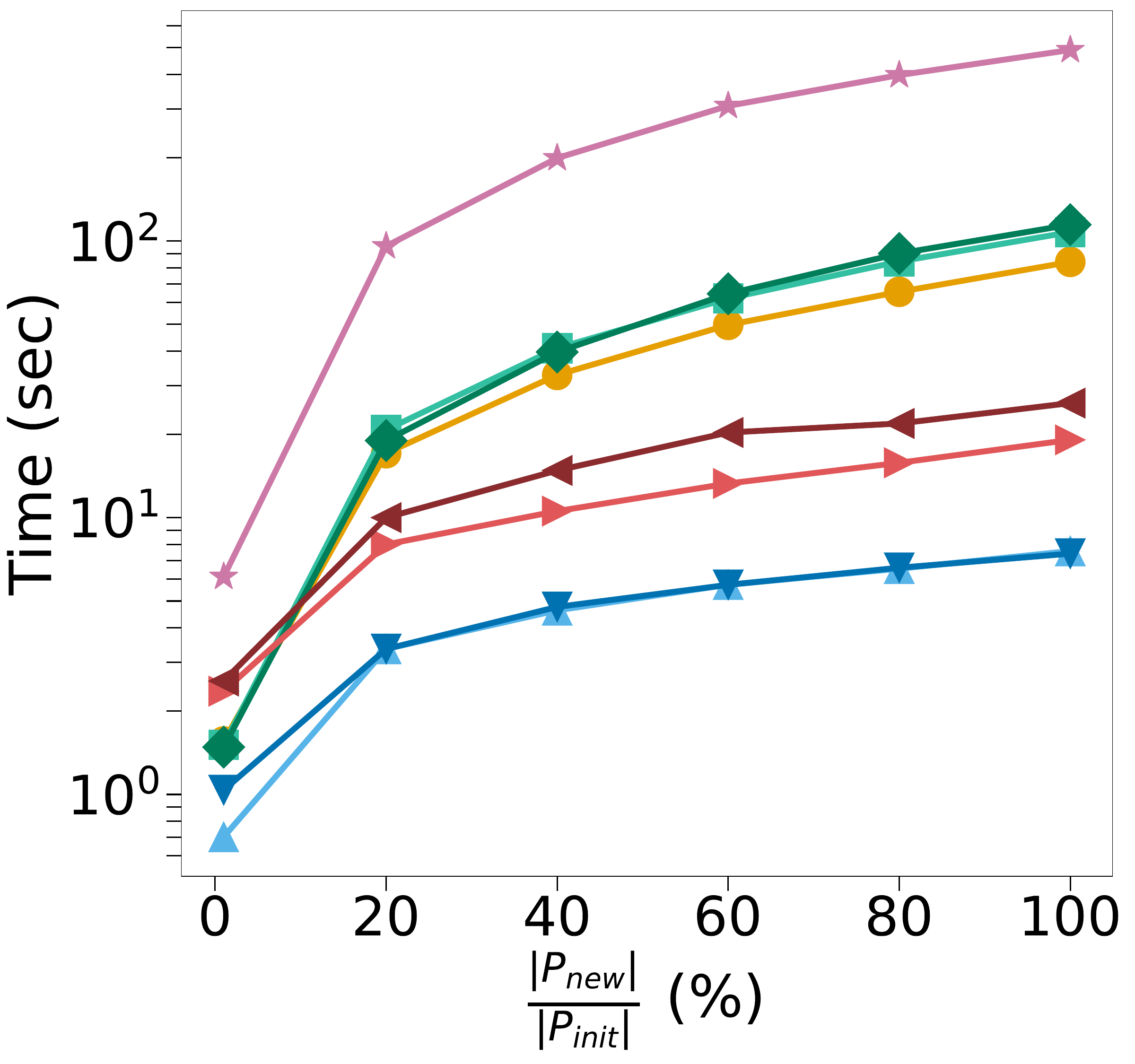}}}}    
    \caption{Runtime of blocking methods}\label{fig:time}
\end{figure}

A primary concern in blocking is ensuring that methods are faster than matching methods and can scale as data size increases. Table~\ref{tab:runtime} and Figure~\ref{fig:time} present the runtime of various blocking methods across datasets of different sizes (measured by the number of entity pairs). The results reflect average runtime over five runs per dataset, with a negligible standard deviation (less than 0.5 seconds) and thus not being reported. Notably, all methods were completed within a reasonable timeframe (less than 1 hour), even for datasets with up to 400 million pairs. Some methods even finished in under a minute for the largest datasets.

While all methods are relatively fast, their scalability varies. This variance can be attributed to the underlying computational processes of each approach. Among the methods tested, \suffix and \exSuffix exhibited the fastest performance and best scalability. These methods generate suffixes from blocking keys, a process known for its linear or near-linear time complexity. Their straightforward operations, like filtering out common suffixes to prevent oversized blocks, enable them to maintain speed even as dataset sizes increase, making them particularly suitable for large-scale applications.

In contrast, \AutoBlock and \CTT, though fast, do not match the performance of the suffix-based methods. These approaches use deep learning models to generate embeddings that group similar records into blocks. Although deep learning can be computationally intensive, \AutoBlock and \CTT are optimized for efficiency. \AutoBlock, for instance, employs an autoencoder—a relatively simple neural network—to produce embeddings quickly. Similarly, \CTT uses a Siamese network to generate embeddings that effectively distinguish between matching and non-matching records. Despite their efficiency, the additional complexity of these methods results in slightly lower performance and scalability compared to the suffix-based methods.

On the other hand, \qgram and \exQgram, which split blocking keys into multiple subsequences, perform faster on smaller datasets but do not scale as well as \AutoBlock and \CTT. The increase in block size and the number of comparisons required as the dataset grows, significantly impacts their efficiency. Finally, \graphblk is the slowest among the tested methods. It involves generating embeddings using a transformer-based model and constructing a graph of database tuples—processes that are computationally intensive. This complexity, especially in the graph clustering stage, leads to longer runtimes and poorer scalability when handling larger datasets.

\starter{Key takeaways} All blocking methods run efficiently, completing within a reasonable timeframe, even for large datasets. The main differences lie in scalability: \suffix and \exSuffix scale best, while \AutoBlock and \CTT are slightly less scalable. \graphblk, \qgram, and \exQgram, struggle with scalability on larger datasets due to their computational complexity.

\begin{table}[h!]
    \centering
    \small
    \resizebox{\linewidth}{!}{
    \begin{tabular}{l@{\hskip 0.03in}c@{\hskip 0.03in}c@{\hskip 0.03in}c@{\hskip 0.03in}c@{\hskip 0.03in}c@{\hskip 0.03in}c@{\hskip 0.03in}c}
    \toprule
    Model &{\scriptsize \amzgog} & {\scriptsize \walamz} & {\scriptsize \DBLPGogS} & {\scriptsize \DBLPACM} & {\scriptsize \Beer} & {\scriptsize \FodorZag} & {\scriptsize \itunamz} \\
    \midrule
    \stdBlock & 4.9s & 1.4m & 3.5m & 8.5s & 3.2s & 0.5s & 22.3m \\
    \qgram & 5.1s & 1.8m & 3.6m & 9.7s & 4.0s & 0.5s & 42.0m \\
    \exQgram & 5.1s & 1.9m & 2.7m & 8.1s & 4.2s & 0.5s & \fbox{51.8m} \\
    \suffix & \textbf{1.5s} & \underline{7.9s} & \textbf{18.9s} & \textbf{2.4s} & \textbf{1.8s} & \textbf{0.3s} & \textbf{9.6s} \\
    \exSuffix & \underline{2.0s} & \textbf{7.8s} & \underline{24.7s} & \underline{4.0s} & \underline{2.4s} & \underline{0.4s} & \underline{15.7s} \\
    \AutoBlock & 5.2s & 18.9s & 58.1s & 7.0s & 9.2s & 0.7s & 1.9m \\
    \CTT & 5.3s & 27.6s & 1.5m & 7.5s & 10.2s & 0.6s & 2.6m \\
    \graphblk & \fbox{58.9s} & \fbox{8.2m} & \fbox{19.2m} & \fbox{2.1m} & \fbox{1.2m} & \fbox{8.2s} & 14.2m \\
    \bottomrule
    \end{tabular}
    }
    \caption{Runtime of blocking methods}
    \vspace{-2mm}
    \label{tab:runtime}
\end{table}

\begin{table}[h!]
    \centering
    \small
    \resizebox{\linewidth}{!}{
    \begin{tabular}{l@{\hskip 0.03in}c@{\hskip 0.03in}c@{\hskip 0.03in}c@{\hskip 0.03in}c@{\hskip 0.03in}c@{\hskip 0.03in}c@{\hskip 0.03in}c}
    \toprule
    Model & {\scriptsize \amzgog} & {\scriptsize \walamz} & {\scriptsize \DBLPGogS} & {\scriptsize \DBLPACM} & {\scriptsize \Beer} & {\scriptsize \FodorZag} & {\scriptsize \itunamz} \\
    \midrule
    \stdBlock & 99.73 & 99.81 & 99.94 & 99.94 & \underline{99.91} & 98.72 & 99.86 \\
    \qgram & 99.69 & 99.77 & 99.95 & 99.94 & 99.90 & 98.83 & 99.79 \\
    \exQgram & 99.70 & 99.75 & 99.95 & 99.94 & 99.90 & 98.88 & 99.77 \\
    \suffix & \underline{99.85} & \underline{99.96} & \underline{99.98} & \underline{99.95} & \textbf{99.95} & \underline{99.31} & \textbf{99.99} \\
    \exSuffix & \textbf{99.86} & \textbf{99.97} & \textbf{99.99} & 99.94 & \textbf{99.95} & \textbf{99.33} & \textbf{99.99} \\
    \AutoBlock & 98.45 & 99.77 & 99.92 & \fbox{97.82} & 98.33 & \fbox{84.89} & \underline{99.91} \\
    \CTT & 98.45 & 99.77 & 99.92 & \fbox{97.82} & 98.33 & \fbox{84.89} & \underline{99.91} \\
    \graphblk & \fbox{97.44} & \fbox{98.47} & \fbox{98.71} & \textbf{99.96} & \fbox{96.71} & 97.31 & \fbox{99.12} \\
    \bottomrule
    \end{tabular}
    }
    \caption{RR values with best (bold), second-best (underlined), and worst (framed) highlighted}
    \label{tab:rr_values}
\end{table}

\subsubsection{Quality of Blocking} \label{sec:exp-blocking}

Table~\ref{tab:rr_values} presents the \RR values for various blocking methods. Most achieve high \RR values close to 1 across datasets, indicating effective candidate pair reduction. However, \suffix and \exSuffix consistently outperform the others, with \exSuffix often achieving the highest \RR. This is due to suffix-based methods' efficiency in creating compact, discriminative blocks by focusing on suffixes and filtering out common ones, avoiding overly large blocks.

In contrast, deep learning methods such as \AutoBlock underperform on smaller datasets like \FodorZag, likely because these methods require larger datasets to generate effective embeddings. A similar trend is observed in midsize datasets like \Beer and \DBLPACM, where deep learning models underperform compared to their performance on larger datasets such as \DBLPGogS and \itunamz. Generally, smaller datasets show lower \RR across all methods due to limited potential for reduction, with this effect being more pronounced in deep learning methods.

\begin{table}[h!]
    \centering
    \small
    \resizebox{\linewidth}{!}{
    \begin{tabular}{l@{\hskip 0.03in}c@{\hskip 0.03in}c@{\hskip 0.03in}c@{\hskip 0.03in}c@{\hskip 0.03in}c@{\hskip 0.03in}c@{\hskip 0.03in}c}
    \toprule
    Model & {\scriptsize \amzgog} & {\scriptsize \walamz} & {\scriptsize \DBLPGogS} & {\scriptsize \DBLPACM} & {\scriptsize \Beer} & {\scriptsize \FodorZag} & {\scriptsize \itunamz} \\
    \midrule
    \stdBlock & \textbf{98.29} & \textbf{99.06} & \underline{98.73} & 99.86 & \textbf{95.59} & \textbf{100.0} & \textbf{97.73} \\
    \qgram & 95.72 & \textbf{99.06} & \textbf{98.75} & \textbf{99.95} & 92.65 & \textbf{100.0} & 73.48 \\
    \exQgram & 94.17 & \underline{98.75} & 97.59 & \textbf{99.95} & 91.18 & \textbf{100.0} & 71.97 \\
    \suffix & 88.52 & 91.16 & 82.38 & \underline{99.91} & 88.24 & \textbf{100.0} & \fbox{50.76} \\
    \exSuffix & \fbox{83.89} & \fbox{88.36} & \fbox{76.47} & 99.46 & 89.71 & \underline{97.32} & 51.52 \\
    \AutoBlock & 88.52 & 96.36 & 95.23 & 99.86 & 85.29 & \textbf{100.0} & 90.15 \\
    \CTT & \underline{95.97} & 97.51 & 95.96 & 99.86 & \underline{94.12} & \textbf{100.0} & \underline{91.67} \\
    \graphblk & 93.92 & 93.66 & 90.03 & \fbox{98.78} & \fbox{83.82} & \fbox{92.86} & 75.00 \\
    \bottomrule
    \end{tabular}
    }
    \caption{\PC values for different models across datasets}
    \label{tab:pc_values}
\end{table}

Table~\ref{tab:pc_values} illustrates \PC. Unlike \RR, where most methods perform well, \PC values show significant performance variations. In datasets such as \amzgog, methods like \exSuffix and \AutoBlock underperform compared to \stdBlock and \CTT. For the \suffix and \exSuffix methods, when suffixes are not distinctive enough, these methods may over-partition the data, splitting true matches into different blocks and reducing \PC. Similarly, in the \itunamz dataset, \qgram, \exQgram, \suffix, and \exSuffix have lower \PC values, missing many true matches. This could be due to the dataset's variability or sparsity, where blocking keys may not align well across records. In contrast, \stdBlock, \AutoBlock, and \CTT preserve more equivalences in \itunamz. \AutoBlock and \CTT, which use embeddings to capture more complex similarities between records, manage to maintain higher \PC even in challenging datasets.

A general takeaway is that suffix-based methods like \suffix and \exSuffix may perform well for datasets with structured and consistent blocking keys. However, deep learning-based methods like \AutoBlock and \CTT, which generate embeddings that can capture complex similarities, may be more effective for datasets with more complex or noisy data. For datasets where the relationships between records are intricate and graph-like, methods like \graphblk could be considered, although their performance may vary depending on how well the embeddings align with the data structure.

%\FodorZag ($6.2\times 10^-4$), \DBLPACM ($3.6 \times 10^{-4}$), \amzgog ($2.7 \times 10^{-4}$), \DBLPGogS ($3.1 \times 10^{-5}$), \walamz ($1.7\times 10^{-5}$),  \Beer ($5.2 \times 10^{-6}$), ,  \itunamz ($3.4 \times 10^{-7}$),  ,

The key dataset property affecting \PC is the ratio of equivalent pairs to the total number of pairs, $|\equivPairs|/|\pairs|$. Intuitively, as this ratio decreases, it becomes more challenging for the blocker to identify true matches among non-matches. For instance, this ratio is approximately $O(10^{-7})$ for \itunamz and $O(10^{-6})$ for \Beer, which have the smallest ratios and thus show the lowest \PC across models. Conversely, \FodorZag has the highest ratio at $O(10^{-4})$ and exhibits the best \PC overall. This pattern generally holds across other datasets.

\begin{table}[htbp]
    \centering
    \small
    \resizebox{\linewidth}{!}{
    \begin{tabular}{l@{\hskip 0.03in}c@{\hskip 0.03in}c@{\hskip 0.03in}c@{\hskip 0.03in}c@{\hskip 0.03in}c@{\hskip 0.03in}c@{\hskip 0.03in}c}
    \toprule
    Model & {\scriptsize \amzgog} & {\scriptsize \walamz} & {\scriptsize \DBLPGogS} & {\scriptsize \DBLPACM} & {\scriptsize \Beer} & {\scriptsize \FodorZag} & {\scriptsize \itunamz} \\
    \midrule
    \stdBlock & 9.58 & 0.90 & 5.34 & \underline{66.26} & 0.54 & 4.96 & 0.02 \\
    \qgram & 8.29 & 0.75 & 6.22 & 59.22 & 0.51 & 5.40 & 0.01 \\
    \exQgram & 8.40 & 0.69 & 5.99 & 60.25 & 0.48 & \underline{5.66} & 0.01 \\
    \suffix & \underline{16.06} & \underline{3.49} & \textbf{16.61} & \textbf{73.06} & \underline{0.88} & \textbf{9.23} & \textbf{0.24} \\
    \exSuffix & \textbf{15.85} & \textbf{4.31} & \underline{17.78} & 60.34 & \textbf{0.89} & \textbf{9.23} & \underline{0.19} \\
    \AutoBlock & 1.52 & 0.73 & 3.89 & \fbox{1.69} & 0.03 & \fbox{0.42} & 0.03 \\
    \CTT & 1.64 & \underline{0.73} & 3.92 & \fbox{1.69} & 0.03 & \fbox{0.42} & 0.04 \\
    \graphblk & \fbox{0.97} & \fbox{0.10} & \fbox{0.22} & 61.75 & \fbox{0.01} & 2.19 & \fbox{0.003} \\
    \bottomrule
    \end{tabular}
    }
    \caption{\PQ values for different models across datasets}
    \label{tab:pq_values}
\end{table}

Table~\ref{tab:pq_values} highlights \PQ's sensitivity to \RR, showing that even a small change in \RR can significantly impact \PQ. For instance, in the \amzgog dataset, \suffix achieves a \PQ of 16.06, while \AutoBlock, with a slightly lower \RR (99.85 for \suffix versus 98.45 for \AutoBlock), sees a drastic drop in \PQ to 1.52. This substantial decline can be attributed to the large number of candidate pairs in \amzgog; when \RR decreases, the number of candidate pairs increases, diluting the concentration of true matches and thereby reducing \PQ. A similar trend appears in \walamz, where \suffix, with a slightly higher \RR, maintains a higher \PQ (3.49) than \exQgram (0.69). In \DBLPGogS, \exSuffix achieves a \PQ of 17.78, outperforming \CTT with the \PQ of 3.92, despite both having relatively high \RR. These examples further show how \PQ is affected by small changes in \RR, particularly in datasets with a large number of pairs.

% Interestingly, the impact of \RR on \PQ is less pronounced in smaller datasets like \Beer. In these datasets, the absolute number of candidate pairs is lower, so variations in \RR have less effect on \PQ. For example, in \itunamz, \PQ values for most methods remain relatively low across the board, with only minor differences. This supports the idea that in smaller datasets, where the pool of candidate pairs is naturally limited, \RR's effect on \PQ is diminished.

\begin{table}[htbp]
    \centering
    \small
    \resizebox{\linewidth}{!}{
    \begin{tabular}{l@{\hskip 0.03in}c@{\hskip 0.03in}c@{\hskip 0.03in}c@{\hskip 0.03in}c@{\hskip 0.03in}c@{\hskip 0.03in}c@{\hskip 0.03in}c}
    \toprule
    Model & {\scriptsize \amzgog} & {\scriptsize \walamz} & {\scriptsize \DBLPGogS} & {\scriptsize \DBLPACM} & {\scriptsize \Beer} & {\scriptsize \FodorZag} & {\scriptsize \itunamz} \\
    \midrule
    \stdBlock & \textbf{99.00} & \textbf{99.44} & \underline{99.33} & 99.90 & \textbf{97.70} & 99.36 & \textbf{98.78} \\
    \qgram & \underline{97.66} & \underline{99.42} & \textbf{99.34} & \textbf{99.95} & 96.14 & 99.41 & 84.64 \\
    \exQgram & 96.86 & 99.25 & 98.75 & \textbf{99.95} & 95.34 & \underline{99.44} & 83.62 \\
    \suffix & 93.84 & 95.36 & 90.33 & \underline{99.93} & 93.73 & \textbf{99.65} & \fbox{67.34} \\
    \exSuffix & \fbox{91.18} & 93.80 & \fbox{86.66} & 99.70 & 94.55 & 98.32 & 68.00 \\
    \AutoBlock & 93.22 & 98.04 & 97.52 & \fbox{98.83} & 91.35 & \fbox{91.83} & 94.78 \\
    \CTT & 97.20 & 98.63 & 97.90 & \fbox{98.83} & \underline{96.18} & \fbox{91.83} & \underline{95.61} \\
    \graphblk & 95.65 & \fbox{96.00} & 94.17 & 99.37 & \fbox{89.81} & 95.03 & 85.41 \\
    \bottomrule
    \end{tabular}
    }
    \caption{\Fb values for different models across datasets}
    \label{tab:fb_values}
\end{table}

As the final quality measure, we report the harmonic mean of \PC and \RR (\Fb) in Table~\ref{tab:fb_values}. The results indicate that no single method consistently outperforms the others across all datasets, reflecting the variability observed in \PC. However, some methods, such as \stdBlock and \qgram, consistently deliver strong performance across various datasets, showcasing stable and reliable results overall. \stdBlock, in particular, achieves the highest or near-highest \Fb values in most cases, making it a robust choice for various datasets. Other methods, like \qgram and \suffix, also perform well in specific datasets, but their effectiveness varies depending on the dataset's properties. For example, \suffix excels in \FodorZag, achieving the highest \Fb, but struggles in \itunamz, where its \Fb drops significantly. This variability highlights the need to select a method aligned with the characteristics of the dataset. In contrast, methods like \AutoBlock and \graphblk tend to have more variable \Fb scores, reflecting their sensitivity to the dataset's complexity and the nature of the data. \AutoBlock, for instance, performs well in some datasets but falls behind in others, particularly in \DBLPACM and \FodorZag. Similarly, \graphblk shows significant fluctuations, with low scores in datasets like \Beer and \itunamz, highlighting its potential limitations in handling certain types of data. Overall, while \stdBlock emerges as a consistently strong performer across datasets, the choice of the optimal blocking method should still be guided by the dataset's specific requirements and characteristics. The variability in \Fb across methods emphasizes the need for careful consideration when selecting a blocking strategy, especially when both high \PC and \RR are critical.

\starter{Key takeaways} In terms of blocking quality, most methods achieve high \RR values, with \suffix and \exSuffix performing best in creating compact and efficient blocks. However, \PC values vary significantly, with methods like \stdBlock and \CTT performing better on datasets with complex relationships, while suffix-based methods may struggle when blocking keys are not sufficiently distinctive. For \Fb, no method consistently outperforms across all datasets. \stdBlock consistently delivers strong performance across most datasets, while \suffix, \AutoBlock, and \graphblk show greater variability, depending on the dataset's complexity and structure.

\subsubsection{Bias Analysis of Blocking Methods} \label{sec:exp-bias}

We now analyze biases in blocking methods. Table~\ref{tab:rr_combined} presents \RR for both minority and majority groups, along with the disparities for each dataset and method. Most disparities are under 1\%, likely because \RR values are close to their maximum of 1, as shown in Table~\ref{tab:rr_values}. Very high \RR values result in consistently high \RR across groups. This is further supported by examining specific methods. Methods such as \suffix and \exSuffix, which have higher \RR values in Table~\ref{tab:rr_values}, consistently show equitable performance across groups in Table~\ref{tab:rr_combined}. In contrast, other methods like \AutoBlock, \CTT, and \graphblk display more variability across datasets, as they exhibit slightly lower \RR values in Table~\ref{tab:rr_values}.

An important observation is that disparities (biases) can be negative, indicating a bias favoring the minority group. This contrasts with the common approach in fairness literature, where disparities are typically reported as absolute values, assuming discrimination against minority groups. In our case, negative disparities may arise from the properties of certain datasets, where minority groups are easier to reduce and partition due to their data distribution. For example, in \DBLPACM dataset, \Beer, \FodorZag, and \itunamz disparities are almost always negative, suggesting better performance for minorities. Conversely, datasets like \amzgog show positive disparities, indicating that blocking is more challenging for minority group, likely due to fewer distinguishing characteristics for this group.

Table~\ref{tab:pc_combined} examines \PC disparities. We first investigate whether higher overall \PC leads to lower \PC bias, as seen with \RR and its disparities. Yet, this relationship does not hold for \PC. For example, \exSuffix shows very small \PC bias in \walamz despite relatively low overall \PC (and \PC per group). Similarly, \exQgram, \suffix, and \exSuffix exhibit lower \PC biases in \itunamz compared to other methods, even though their overall \PC performance in this dataset is subpar. The reverse can also occur; \stdBlock achieves the best overall \PC in \itunamz but shows significant \PC bias (7.5). This differs from \RR because overall \PC is not close enough to 1, leaving room for disparity between \PC values for different groups. Thus, even with strong overall performance, bias can persist. This observation highlights an important point: improving overall \PC and blocking quality may not necessarily reduce biases and disparities. Specialized methods are required to address bias reduction in \PC as a primary objective.

Table~\ref{tab:fb_combined} highlights \Fb disparities across methods and datasets. Similar to what we observed in Table~\ref{tab:pc_values} and~\ref{tab:fb_values}, \Fb mainly mirrors \PC, also in their biases. This, again, can be explained by the fact that \RR is often close to 1, and its biases are insignificant. Therefore, biases in \Fb mainly reflect those in \PC.

\starter{Key takeaways} Most blocking methods exhibit low \RR bias due to their consistently high \RR values, making disparities in \RR insignificant. However, \PC biases vary more across methods and datasets, with deep learning-based methods showing greater disparity in some cases. Negative disparities are observed in several datasets, favoring minority groups, likely due to dataset characteristics. Importantly, improving overall blocking quality does not necessarily reduce bias, and addressing bias may require specialized techniques beyond simply enhancing \PC or \RR performance.

\begin{table*}[htbp]
    \centering
    \small
    \resizebox{\textwidth}{!}{
    \begin{tabular}{l c c c c c c c}
    \toprule
    \textbf{Model} & \amzgog & \walamz & \DBLPGogS & \DBLPACM & \Beer & \FodorZag & \itunamz \\
    \midrule

    \stdBlock & 
0.07 (99.73, 99.66) & 
0.14 (99.82, 99.68) & 
\fbox{-0.05} (99.94, 99.99) & 
\textbf{-0.01} (99.94, 99.95) & 
\underline{-0.01} (99.90, 99.91) & 
\underline{0.01} (98.72, 98.71) & 
-0.13 (99.80, 99.93) \\

\qgram & 
0.39 (99.72, 99.33) & 
0.25 (99.79, 99.54) & 
-0.03 (99.95, 99.98) & 
\underline{-0.02} (99.93, 99.95) & 
\underline{-0.01} (99.90, 99.91) & 
-0.20 (98.80, 99.00) & 
-0.15 (99.73, 99.88) \\

\exQgram & 
0.41 (99.73, 99.32) & 
0.18 (99.76, 99.58) & 
-0.02 (99.95, 99.97) & 
\textbf{-0.01} (99.94, 99.95) & 
\textbf{0.00} (99.90, 99.90) & 
-0.15 (98.86, 99.01) & 
-0.20 (99.68, 99.88) \\

\suffix & 
0.07 (99.86, 99.79) & 
-0.02 (99.95, 99.97) & 
\textbf{0.00} (99.98, 99.98) & 
\textbf{-0.01} (99.95, 99.96) & 
-0.02 (99.94, 99.96) & 
-0.18 (99.28, 99.46) & 
\textbf{0.00} (99.99, 99.99) \\

\exSuffix & 
\underline{0.04} (99.86, 99.82) & 
\underline{-0.01} (99.96, 99.97) & 
\textbf{0.00} (99.99, 99.99) & 
\textbf{-0.01} (99.94, 99.95) & 
-0.02 (99.94, 99.96) & 
-0.18 (99.30, 99.48) & 
\textbf{0.00} (99.99, 99.99) \\

\AutoBlock & 
\textbf{-0.01} (98.45, 98.46) & 
-0.02 (99.77, 99.79) & 
\underline{0.01} (99.92, 99.91) & 
\fbox{-0.48} (97.73, 98.21) & 
0.05 (98.36, 98.31) & 
\fbox{0.90} (85.02, 84.12) & 
\underline{-0.04} (99.89, 99.93) \\

\CTT & 
\textbf{-0.01} (98.45, 98.46) & 
\textbf{0.00} (99.77, 99.77) & 
\textbf{0.00} (99.92, 99.92) & 
-0.21 (97.78, 97.99) & 
-0.09 (98.29, 98.38) & 
\textbf{0.00} (84.89, 84.89) & 
\underline{-0.04} (99.89, 99.93) \\

\graphblk & 
\fbox{0.72} (97.49,96.77) & 
\fbox{0.43} (98.49,98.06)& 
\fbox{0.05} (98.71, 98.66)& 
\underline{-0.02} (99.95, 99.97) &
\fbox{-0.81} (96.29,97.10 ) & 
0.15 (97.33  , 97.18)& 
\fbox{0.25} (99.28  , 99.03) \\

    \bottomrule
    \end{tabular}
    }
    \caption{For each cell, a (b,c) shows \RR values within minority and majority groups (b,c) and the \RR disparities (a)}
    \label{tab:rr_combined}
\end{table*}

\begin{table*}[htbp]
    \centering
    \small
    \resizebox{\textwidth}{!}{
    \begin{tabular}{l c c c c c c c}
    \toprule
    \textbf{Model} & \amzgog & \walamz & \DBLPGogS & \DBLPACM & \Beer & \FodorZag & \itunamz \\
    \midrule

\stdBlock & 
1.70 (98.37, 96.67) & 
1.47 (99.20, 97.73) & 
0.78 (98.79, 98.01) & 
-0.16 (99.84, 100.00) & 
\underline{-1.68} (94.87, 96.55) & 
\textbf{0.00} (100.00, 100.00) & 
7.50 (100.00, 92.50) \\

\qgram & 
\underline{-1.01} (95.66, 96.67) & 
1.47 (99.20, 97.73) & 
1.33 (98.85, 97.52) & 
\textbf{-0.05} (99.95, 100.00) & 
-6.81 (89.74, 96.55) & 
\textbf{0.00} (100.00, 100.00) & 
\fbox{-9.35} (70.65, 80.00) \\

\exQgram & 
6.16 (94.49, 88.33) & 
\underline{1.13} (98.86, 97.73) & 
1.14 (97.67, 96.53) & 
\textbf{-0.05} (99.95, 100.00) & 
\fbox{-9.37} (87.18, 96.55) & 
\textbf{0.00} (100.00, 100.00) & 
\textbf{-0.76} (71.74, 72.50) \\

\suffix & 
16.01 (89.34, 73.33) & 
5.29 (91.65, 86.36) & 
\textbf{0.27} (82.40, 82.13) & 
\underline{-0.10} (99.90, 100.00) & 
-8.48 (84.62, 93.10) & 
\textbf{0.00} (100.00, 100.00) & 
\underline{1.09} (51.09, 50.00) \\

\exSuffix & 
\fbox{18.15} (84.82, 66.67) & 
\textbf{0.94} (88.44, 87.50) & 
1.66 (76.60, 74.94) & 
-0.63 (99.37, 100.00) & 
-5.92 (87.18, 93.10) & 
\underline{-2.97} (97.03, 100.00) & 
2.17 (52.17, 50.00) \\

\AutoBlock & 
8.98 (88.98, 80.00) & 
4.75 (96.80, 92.05) & 
-0.59 (95.19, 95.78) & 
-0.16 (99.84, 100.00) & 
\textbf{-1.59} (84.62, 86.21) & 
\textbf{0.00} (100.00, 100.00) & 
-6.96 (88.04, 95.00) \\

\CTT & 
2.79 (96.12, 93.33) & 
4.76 (97.94, 93.18) & 
\underline{0.47} (96.00, 95.53) & 
-0.16 (99.84, 100.00) & 
-4.24 (92.31, 96.55) & 
\textbf{0.00} (100.00, 100.00) & 
-1.20 (91.30, 92.50) \\

\graphblk & 
\textbf{0.62} (93.95 , 93.33) & 
\fbox{5.52} (94.16 , 88.64) & 
\fbox{-3.26} (89.79 , 93.05) & 
\fbox{-1.04} (98.64, 99.68) & 
7.87 (87.18 , 79.31) & 
\fbox{-7.92} (92.08 , 100.0) & 
7.17 (77.17 , 70.00) \\

    \bottomrule
    \end{tabular}
    }
    \caption{\PC disparities and \PC values within the minority and majority groups}
    \label{tab:pc_combined}
\end{table*}

\begin{table*}[htbp]
    \centering
    \small
    \resizebox{\textwidth}{!}{
    \begin{tabular}{l c c c c c c c}
    \toprule
    \textbf{Model} & \amzgog & \walamz & \DBLPGogS & \DBLPACM & \Beer & \FodorZag & \itunamz \\
    \midrule
\stdBlock & 
0.91 (99.05, 98.14) & 
0.82 (99.51, 98.69) & 
0.37 (99.36, 98.99) & 
-0.08 (99.89, 99.97) & 
\textbf{-0.88} (97.32, 98.20) & 
\textbf{0.01} (99.36, 99.35) & 
3.83 (99.90, 96.07) \\

\qgram & 
\textbf{-0.33} (97.65, 97.98) & 
0.87 (99.49, 98.62) & 
0.66 (99.39, 98.73) & 
\textbf{-0.03} (99.94, 99.97) & 
-3.65 (94.55, 98.20) & 
-0.10 (99.39, 99.49) & 
\fbox{-6.13} (82.71, 88.84) \\

\exQgram & 
3.53 (97.04, 93.51) & 
\underline{0.66} (99.31, 98.65) & 
0.58 (98.80, 98.22) & 
\textbf{-0.03} (99.94, 99.97) & 
\fbox{-5.09} (93.11, 98.20) & 
-0.07 (99.42, 99.49) & 
\textbf{-0.59} (83.43, 84.02) \\

\suffix & 
9.77 (94.31, 84.54) & 
2.95 (95.62, 92.67) & 
\textbf{0.17} (90.35, 90.18) & 
\underline{-0.06} (99.92, 99.98) & 
-4.77 (91.64, 96.41) & 
-0.09 (99.64, 99.73) & 
0.95 (67.62, 66.67) \\

\exSuffix & 
\fbox{11.79} (91.73, 79.94) & 
\textbf{0.53} (93.85, 93.32) & 
1.07 (86.74, 85.67) & 
-0.32 (99.65, 99.97) & 
-3.29 (93.12, 96.41) & 
-1.59 (98.15,99.74) & 
1.91 (68.57, 66.66) \\

\AutoBlock & 
5.21 (93.48, 88.27) & 
2.50 (98.26, 95.76) & 
-0.30 (97.50, 97.80) & 
-0.32 (98.78, 99.10) & 
\underline{-0.89} (90.97, 91.86) & 
0.58 (91.91, 91.33) & 
-3.81 (93.59, 97.40) \\

\CTT & 
1.44 (97.27, 95.83) & 
2.49 (98.85, 96.36) & 
\underline{0.24} (97.92, 97.68) & 
-0.19 (98.80, 98.99) & 
-2.26 (95.20, 97.46) & 
\underline{0.02} (91.83, 91.81) & 
\underline{-0.66} (95.41, 96.07) \\

\graphblk & 
\underline{0.65} (95.67 , 95.02)&
\fbox{3.17} (96.28, 93.11)&
\fbox{-1.73} (94.04, 95.77)&
\fbox{-0.53} (99.29, 99.82) & 
4.20 (91.51, 87.31)&
\fbox{-3.56} (94.63, 98.19)&
4.82 (86.84, 82.02) \\

    \bottomrule
    \end{tabular}
    }
    \caption{\Fb disparities and \Fb values within the minority and majority groups}
    \label{tab:fb_combined}
\end{table*}

\subsubsection{Assessing Bias Propagation from Blocking to Matching} \label{sec:bias-impact}

To demonstrate the significance of bias in blocking, we conduct an experiment to show how this bias translates into bias in the matching results. To assess the bias introduced by the blocking stage in an end-to-end EM pipeline, we focus solely on the bias stemming from the blocker, excluding the matcher. We assume a hypothetical perfect matcher that produces no false positives or false negatives. This allows us to isolate the effect of bias in the blocker on the final outcomes without introducing bias from the matcher. Although a perfect matcher might seem unnecessary with a blocker, we assume that the matcher is computationally expensive and can only process a limited number of pairs, making the blocker essential.

For the experiment, we select a dataset and, for each blocker, generate a candidate set representing the output of that blocker. These candidate pairs are processed by the perfect matcher, and we assess fairness using Equal Opportunity Difference (\EOD), Equalized Odds (\EO), and Demographic Parity (\DP). \EO measures the difference in True Positive Rates between groups. \EOD sums the differences in True Positive Rates and False Positive Rates across groups. \DP measures the difference in groups' prediction rates, regardless of true labels~\cite{hardt2016equality}.

We present results for \amzgog, comparing \qgram (with the lowest \PC disparity) and \exSuffix (with the highest \PC disparity). Results for other datasets and methods are similar and are available in our GitHub repository. Table~\ref{tab:comparison_metrics} shows the outcomes. \exSuffix had a \PC disparity of 18.15\%, while \qgram had a much lower \PC disparity of 1.01\% (absolute values). As it is shown in Table~\ref{tab:comparison_metrics}, our hypothesis that blocker bias would carry over to the final EM results is confirmed. \exSuffix, with higher \PC disparity, showed fairness metrics of 18.16\% for both \EO and \EOD, indicating significantly greater bias. In contrast, \qgram, with much lower \PC disparity, had fairness metrics of only 1.01\% for \EO and \EOD, indicating much less bias. \qgram also had much less \DP disparity compared to \exSuffix, further confirming our claim. This experiment highlights that blocker bias significantly affects the overall bias in the EM pipeline.

% \starter{Key takeaways} The experiment shows that bias introduced in the blocking stage can propagate to the matching results, even with a perfect matcher. Blockers with higher \PC disparities, like \exSuffix, lead to significantly larger fairness metric values, while those with lower disparities, like \qgram, show much less bias. This confirms that bias in the blocker can have a substantial impact on the fairness of the overall EM process.

\starter{Key takeaways} Bias introduced by the blocking stage can propagate to the matching results, even when using a perfect matcher. Blockers with higher \PC disparities, like \exSuffix, result in significantly larger fairness metric values, while those with lower disparities, like \qgram, show much less bias. This confirms that bias in the
blocker can have a substantial impact on the fairness of the overall EM process.

\begin{table}[htbp]
    \centering
    \large
    \resizebox{\linewidth}{!}{
    \begin{tabular}{l@{\hskip 0.08in}c@{\hskip 0.08in}c@{\hskip 0.08in}c@{\hskip 0.08in}c@{\hskip 0.08in}c}
    \toprule
    \multirow{2}{*}{\textbf{Metric}} & \multirow{2}{*}{\DP (\%)} & \multirow{2}{*}{\EO (\%)} & \multirow{2}{*}{\EOD (\%)} & \multicolumn{2}{c}{{(TP, FP, TN, FN)}} \\
    \cmidrule(lr){5-6}
     & & & & Minority & Majority \\
    \midrule
    \qgram     & 4.42$\times 10^{-3}$ & 1.01 & 1.01 & (40, 0, 272818, 20) & (939, 0, 4123053, 168) \\
    \exSuffix  & 8.11$\times 10^{-3}$ & 18.16 & 18.16 & (58, 0, 272818, 2)  & (1059, 0, 4123053, 48) \\
    \bottomrule
    \end{tabular}
    }
    \caption{Fairness metrics comparison for \qgram and \exSuffix on \amzgog}
    \label{tab:comparison_metrics}
\end{table}

\subsubsection{Impact of Removing Sensitive Attributes} \label{sec:exp-nosens}

As a first step toward mitigating biases, we conducted an experiment to test ``fairness through unawareness'' in the context of blocking. The idea is that an algorithm may become fairer if it avoids using sensitive attributes like race or gender in its decision-making process. However, this approach has limitations, as the algorithm may still act on correlated attributes, resulting in biased outcomes. The goal of this experiment is to evaluate this hypothesis~\cite{dwork2012fairness}.

\begin{figure}[htb]
   \centering
   \subfloat[\suffix]{\label{fig:std_exp3}
			{{\includegraphics[width=0.23\textwidth]{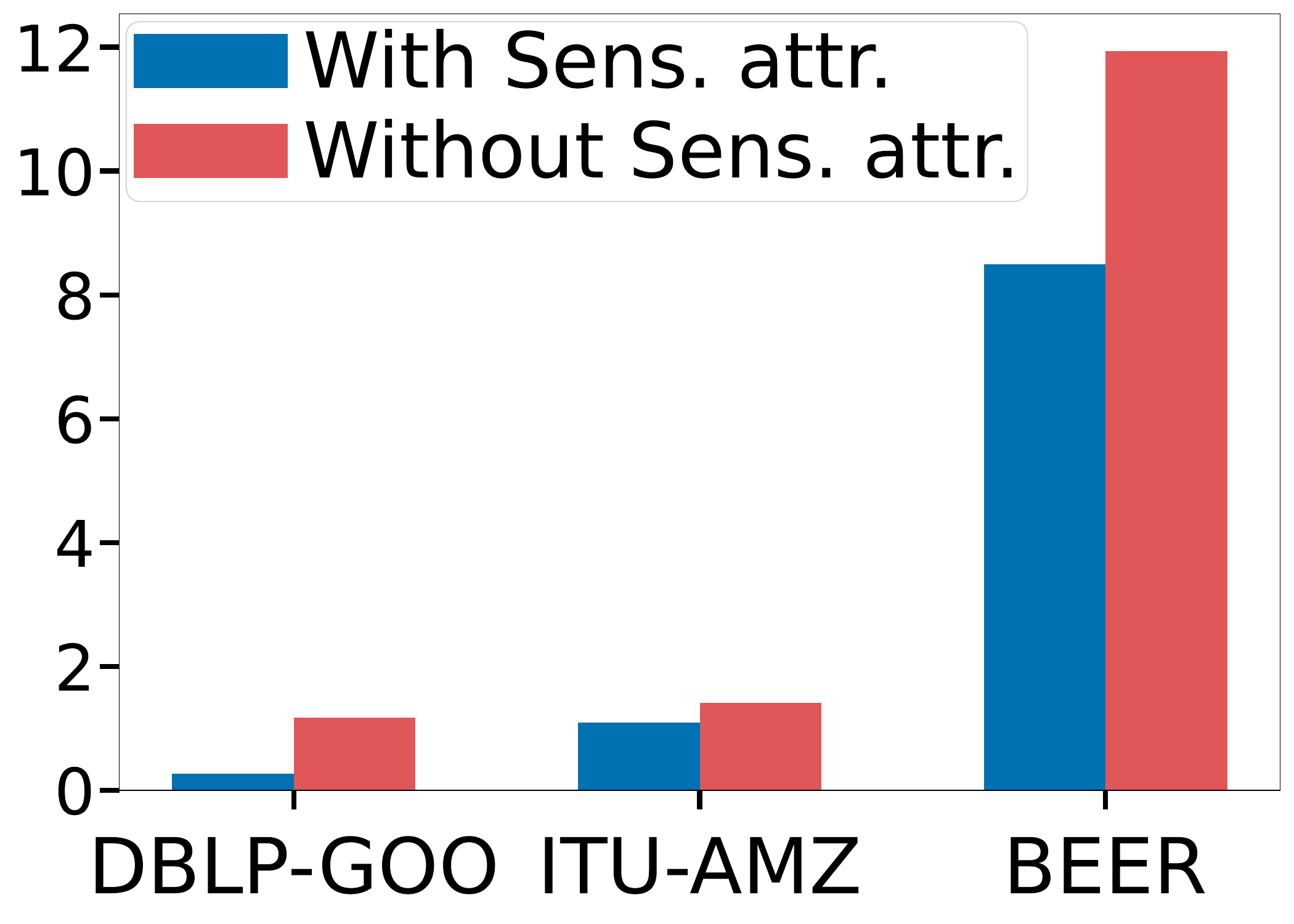}}}}	
   \subfloat[\AutoBlock]{\label{fig:qgram_exp3}   
    {{\includegraphics[width=0.23\textwidth]{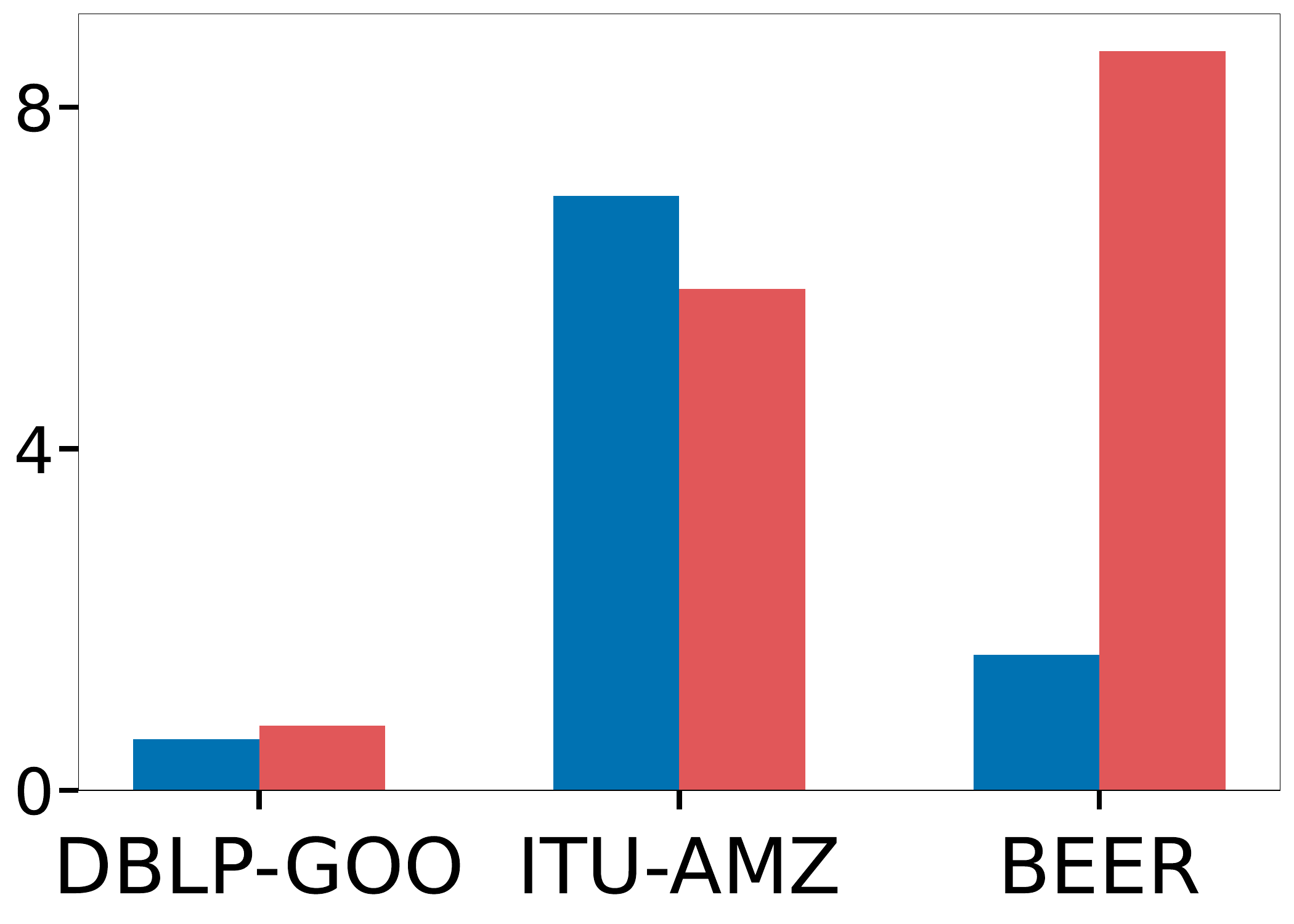}}}}        \caption{Impact of removing sensitives on methods}\label{fig:noSensMethod}
\end{figure}

\begin{figure}[htb]
   \centering
   \subfloat[\walamz]{\label{fig:qgram_exp31}   
    {{\includegraphics[width=0.23\textwidth]{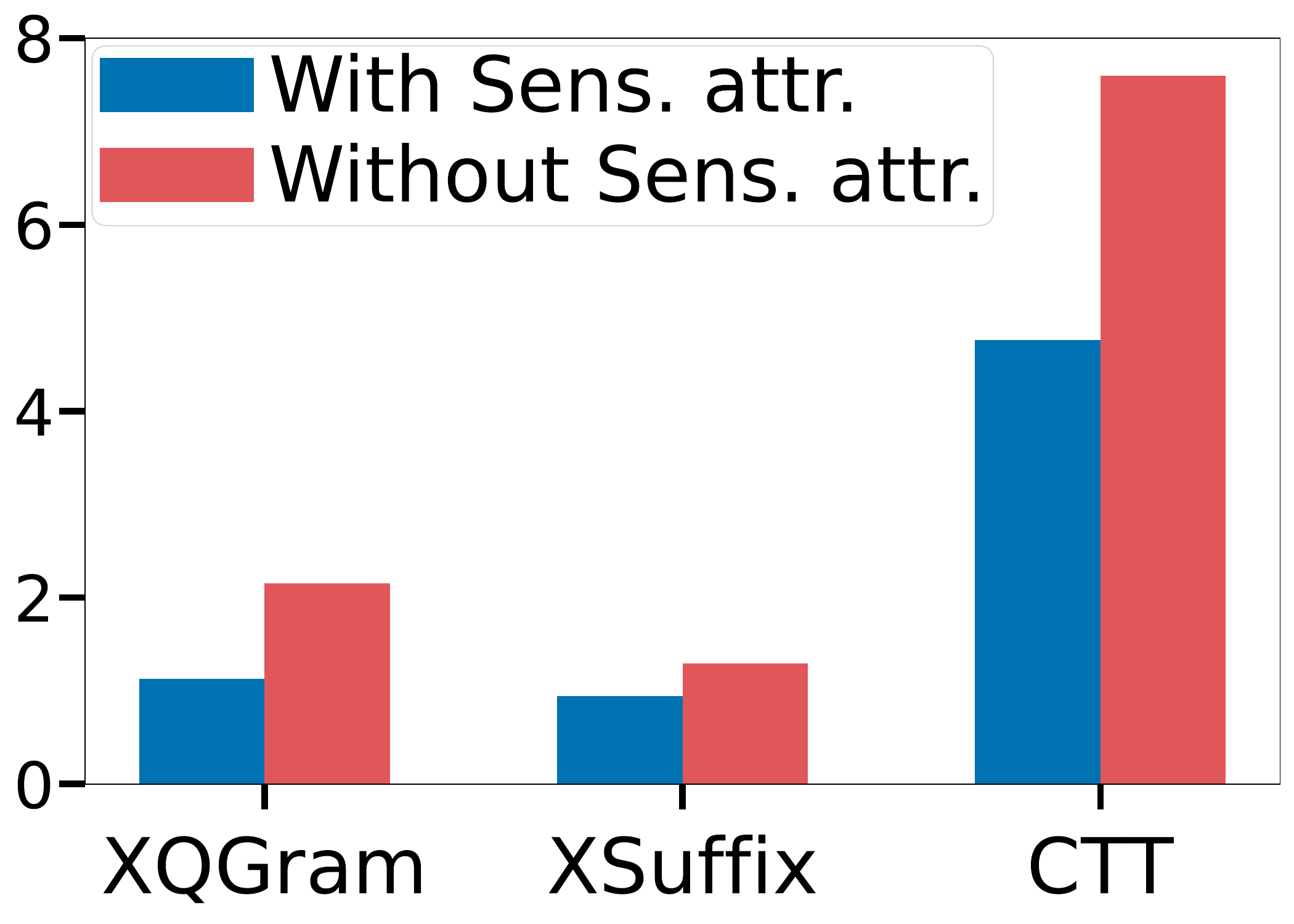}}}}      
       \subfloat[\amzgog]{\label{fig:std_exp31}
			{{\includegraphics[width=0.23\textwidth]{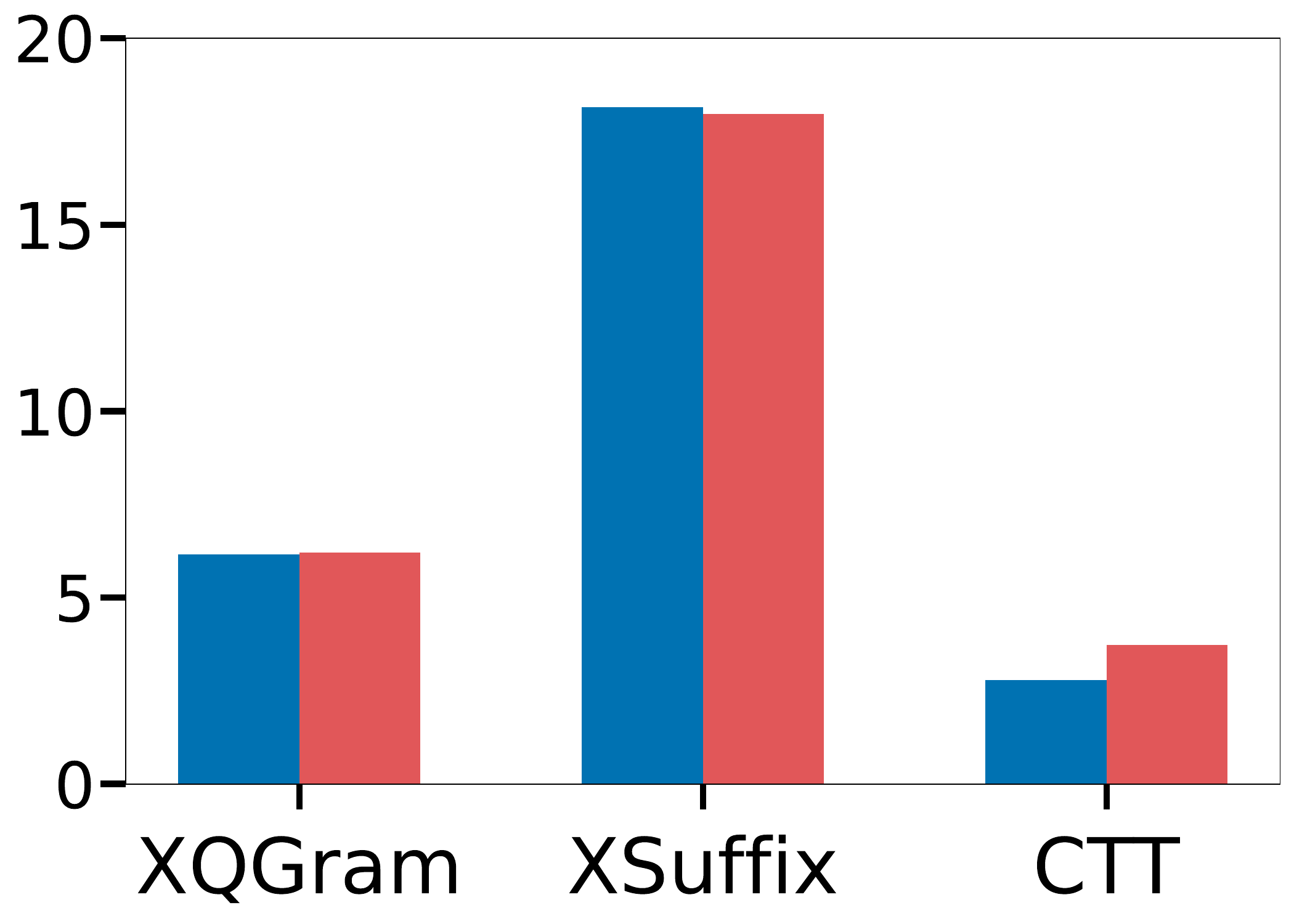}}}}
    \caption{Impact of removing sensitives in datasets}\label{fig:NoSesnsData}
\end{figure}

Figures ~\ref{fig:noSensMethod} and ~\ref{fig:NoSesnsData} demonstrate the \PC disparity in two scenarios: first, when blocking is performed using all attributes in the dataset, including the sensitive attribute, and second, when blocking is done using all attributes except the sensitive one. We present a few comparisons in this paper, but the complete set of comparisons is available in our GitHub repository.

In most cases, we observe an increase in disparity. The reasoning behind this is that when the sensitive attribute is removed, the algorithm may still rely on other correlated attributes that indirectly capture sensitive attributes information. However, without the sensitive attribute explicitly guiding the process, the blocking may become less effective at capturing the distinctive features of minority groups, resulting in increased disparity.

In some instances, there is little to no change in disparity. This suggests that the remaining non-sensitive attributes are sufficient to create balanced blocking. This can happen when the sensitive attribute is not strongly correlated with other key attributes used in blocking, or when the dataset contains other robust features that dominate the blocking process.

Interestingly, in cases like \AutoBlock on \itunamz, removing the sensitive attribute can slightly reduce bias. This occurs when the sensitive attribute unintentionally introduces bias into the blocking process, such as when it leads to the over-blocking of a particular group. Removing the sensitive attribute can, in such cases, lead to more balanced blocking.

\starter{Key takeaways} Overall, this experiment highlights a key limitation of the ``fairness through unawareness'' approach: {\it excluding sensitive attributes does not necessarily prevent biased outcomes}, as other correlated features may still result in similar patterns of disparity. In most cases, removing the sensitive attribute increased bias, reinforcing the concern that ignoring sensitive attributes is not a guaranteed solution to mitigating bias in blocking methods.

\section{Conclusion and Future Work}\label{sec:conclusion}

This paper evaluated blocking methods in EM and their associated biases. Our experiments demonstrated how {\bf blocking biases propagate into final EM outcomes}, showing that while blocking reduces computational complexity, it can {\bf introduce significant biases, particularly when sensitive or correlated attributes are involved}. These biases often carry out to the final matching results, with some blocking methods exacerbating them more than others. There was no single blocking method that consistently demonstrated lower disparities across all datasets. Instead, the best-performing method in terms of fairness varied depending on the characteristics of each dataset. Our experiment on removing sensitive attributes highlights the limitations of ``fairness through unawareness,'' as bias persisted due to the influence of correlated features.

There are several directions for future research. First, developing {\bf debiasing methods specifically tailored to blocking} is essential. This could involve improving existing algorithms or introducing new ones that consistently reduce bias while maintaining performance across datasets and domains. Second, future research should explore bias evaluation through the lens of intersectionality, considering multiple sensitive attributes (e.g., gender and ethnicity) simultaneously. Finally, debiasing techniques should extend beyond blocking to the entire end-to-end EM pipeline, covering both blocking and matching stages.

% This paper contributes to ongoing efforts in the research community to improve fairness in data preparation and cleaning. Fairness in EM is crucial for building responsible AI systems. As part of this contribution, we have released the implementation of our experiments to support further research on fairness and bias mitigation in EM.

This paper contributes to ongoing efforts to improve fairness in data preparation and cleaning, which is crucial for building responsible AI systems. To support further research on fairness and bias mitigation in EM, we have released the implementation of our experiments.

\bibliographystyle{IEEEtran}
\bibliography{IEEEabrv,references}

\begin{thebibliography}{10}
\providecommand{\url}[1]{#1}
\csname url@samestyle\endcsname
\providecommand{\newblock}{\relax}
\providecommand{\bibinfo}[2]{#2}
\providecommand{\BIBentrySTDinterwordspacing}{\spaceskip=0pt\relax}
\providecommand{\BIBentryALTinterwordstretchfactor}{4}
\providecommand{\BIBentryALTinterwordspacing}{\spaceskip=\fontdimen2\font plus
\BIBentryALTinterwordstretchfactor\fontdimen3\font minus \fontdimen4\font\relax}
\providecommand{\BIBforeignlanguage}[2]{{%
\expandafter\ifx\csname l@#1\endcsname\relax
\typeout{** WARNING: IEEEtran.bst: No hyphenation pattern has been}%
\typeout{** loaded for the language `#1'. Using the pattern for}%
\typeout{** the default language instead.}%
\else
\language=\csname l@#1\endcsname
\fi
#2}}
\providecommand{\BIBdecl}{\relax}
\BIBdecl

\bibitem{mudgal2018deep}
S.~Mudgal, H.~Li, T.~Rekatsinas, A.~Doan, Y.~Park, G.~Krishnan, R.~Deep, E.~Arcaute, and V.~Raghavendra, ``Deep learning for entity matching: A design space exploration,'' in \emph{SIGMOD}, 2018.

\bibitem{fu2021hierarchical}
C.~Fu, X.~Han, J.~He, and L.~Sun, ``Hierarchical matching network for heterogeneous entity resolution,'' in \emph{IJCAI}, 2021.

\bibitem{li2020deep}
Y.~Li, J.~Li, Y.~Suhara, A.~Doan, and W.-C. Tan, ``Deep entity matching with pre-trained language models,'' \emph{PVLDB}, 2020.

\bibitem{yao2022entity}
D.~Yao, Y.~Gu, G.~Cong, H.~Jin, and X.~Lv, ``Entity resolution with hierarchical graph attention networks,'' in \emph{SIGMOD}, 2022.

\bibitem{konda2018magellan}
P.~Konda, S.~Das, A.~Doan, A.~Ardalan, J.~R. Ballard, H.~Li, F.~Panahi, H.~Zhang, J.~Naughton, S.~Prasad \emph{et~al.}, ``Magellan: toward building entity matching management systems over data science stacks,'' \emph{PVLDB}, 2016.

\bibitem{michelson2006learning}
M.~Michelson and C.~A. Knoblock, ``Learning blocking schemes for record linkage,'' in \emph{AAAI}, 2006.

\bibitem{ebraheem2018distributed}
M.~Ebraheem, S.~Thirumuruganathan, S.~Joty, M.~Ouzzani, and N.~Tang, ``Distributed representations of tuples for entity resolution,'' \emph{VLDB}, 2018.

\bibitem{papadakis2020blocking}
G.~Papadakis, D.~Skoutas, E.~Thanos, and T.~Palpanas, ``Blocking and filtering techniques for entity resolution: A survey,'' \emph{CSUR}, 2020.

\bibitem{li2020survey}
B.-H. Li, Y.~Liu, A.-M. Zhang, W.-H. Wang, and S.~Wan, ``A survey on blocking technology of entity resolution,'' \emph{JCS\&T}, 2020.

\bibitem{mccallum2000efficient}
A.~McCallum, K.~Nigam, and L.~H. Ungar, ``Efficient clustering of high-dimensional data sets with application to reference matching,'' in \emph{SIGKDD}, 2000.

\bibitem{bgp}
L.~O. Evangelista, E.~Cortez, A.~S. da~Silva, and W.~Meira~Jr, ``Adaptive and flexible blocking for record linkage tasks,'' \emph{JIDM}, 2010.

\bibitem{zhang2020autoblock}
W.~Zhang, H.~Wei, B.~Sisman, X.~L. Dong, C.~Faloutsos, and D.~Page, ``Autoblock: A hands-off blocking framework for entity matching,'' in \emph{WSDM}, 2020.

\bibitem{zafar2017fairness}
M.~B. Zafar, I.~Valera, M.~Gomez~Rodriguez, and K.~P. Gummadi, ``Fairness beyond disparate treatment \& disparate impact: Learning classification without disparate mistreatment,'' in \emph{WWW}, 2017.

\bibitem{hardt2016equality}
M.~Hardt, E.~Price, and N.~Srebro, ``Equality of opportunity in supervised learning,'' in \emph{NIPS}, 2016.

\bibitem{dwork2012fairness}
C.~Dwork, M.~Hardt, T.~Pitassi, O.~Reingold, and R.~Zemel, ``Fairness through awareness,'' in \emph{ITCS}, 2012.

\bibitem{efthymiou2021fairer}
V.~Efthymiou, K.~Stefanidis, E.~Pitoura, and V.~Christophides, ``{FairER:} entity resolution with fairness constraints,'' in \emph{CIKM}, 2021.

\bibitem{nilforoushan2022entity}
S.~Nilforoushan, Q.~Wu, and M.~Milani, ``Entity matching with auc-based fairness,'' in \emph{IEEE Big Data}, 2022.

\bibitem{shahbazi2023through}
N.~Shahbazi, N.~Danevski, F.~Nargesian, A.~Asudeh, and D.~Srivastava, ``Through the fairness lens: Experimental analysis and evaluation of entity matching,'' \emph{VLDB}, 2023.

\bibitem{moslemi2024threshold}
M.~H. Moslemi and M.~Milani, ``Threshold-independent fair matching through score calibration,'' in \emph{GUIDE-AI at SIGMOD}, 2024.

\bibitem{shahbazi2024fairness}
N.~Shahbazi, J.~Wang, Z.~Miao, and N.~Bhutani, ``Fairness-aware data preparation for entity matching,'' in \emph{ICDE}.\hskip 1em plus 0.5em minus 0.4em\relax IEEE, 2024.

\bibitem{papadakis2016comparative}
G.~Papadakis, J.~Svirsky, A.~Gal, and T.~Palpanas, ``Comparative analysis of approximate blocking techniques for entity resolution,'' \emph{PVLDB}, 2016.

\bibitem{christen2011survey}
P.~Christen, ``A survey of indexing techniques for scalable record linkage and deduplication,'' \emph{TKDE}, 2011.

\bibitem{fellegi1969theory}
I.~P. Fellegi and A.~B. Sunter, ``A theory for record linkage,'' \emph{JASA}, 1969.

\bibitem{hernandez1995sorted}
M.~A. Hern{\'a}ndez and S.~J. Stolfo, ``The sorted neighborhood method for duplicate record detection,'' in \emph{SIGMOD}, 1995.

\bibitem{sarma2011cblock}
A.~Das~Sarma, A.~Jain, A.~Machanavajjhala, and P.~Bohannon, ``An automatic blocking mechanism for large-scale de-duplication tasks,'' in \emph{CIKM}, 2012.

\bibitem{javdani2019deepblock}
D.~Javdani, H.~Rahmani, M.~Allahgholi, and F.~Karimkhani, ``Deepblock: A novel blocking approach for entity resolution using deep learning,'' in \emph{ICWR}, 2019.

\bibitem{christen2007quality}
P.~Christen and K.~Goiser, ``Quality and complexity measures for data linkage and deduplication,'' in \emph{Quality measures in data mining}.\hskip 1em plus 0.5em minus 0.4em\relax Springer, 2007.

\bibitem{elfeky2002tailor}
M.~G. Elfeky, V.~S. Verykios, and A.~K. Elmagarmid, ``Tailor: A record linkage toolbox,'' in \emph{ICDE}, 2002.

\bibitem{thirumuruganathan2021deep}
S.~Thirumuruganathan, H.~Li, N.~Tang, M.~Ouzzani, Y.~Govind, D.~Paulsen, G.~Fung, and A.~Doan, ``Deep learning for blocking in entity matching: a design space exploration,'' \emph{PVLDB}, 2021.

\bibitem{papadakis2011efficient}
G.~Papadakis, E.~Ioannou, C.~Nieder{\'e}e, and P.~Fankhauser, ``Efficient entity resolution for large heterogeneous information spaces,'' in \emph{WSDM}, 2011.

\bibitem{steorts2014comparison}
R.~C. Steorts, S.~L. Ventura, M.~Sadinle, and S.~E. Fienberg, ``A comparison of blocking methods for record linkage,'' in \emph{PSD}, 2014.

\bibitem{de2009robust}
T.~De~Vries, H.~Ke, S.~Chawla, and P.~Christen, ``Robust record linkage blocking using suffix arrays,'' in \emph{CIKM}, 2009.

\bibitem{galhotra2021efficient}
S.~Galhotra, D.~Firmani, B.~Saha, and D.~Srivastava, ``Efficient and effective er with progressive blocking,'' \emph{VLDB}, 2021.

\bibitem{zeakis2023pre}
A.~Zeakis, G.~Papadakis, D.~Skoutas, and M.~Koubarakis, ``Pre-trained embeddings for entity resolution: an experimental analysis,'' \emph{PVLDB}, 2023.

\bibitem{o2019review}
K.~O’Hare, A.~Jurek-Loughrey, and C.~d. Campos, ``A review of unsupervised and semi-supervised blocking methods for record linkage,'' \emph{Linking and Mining Heterogeneous and Multi-view Data}, 2019.

\bibitem{o2018new}
K.~O’Hare, A.~Jurek, and C.~de~Campos, ``A new technique of selecting an optimal blocking method for better record linkage,'' \emph{Information Systems}, 2018.

\bibitem{kopcke2010frameworks}
H.~K{\"o}pcke and E.~Rahm, ``Frameworks for entity matching: A comparison,'' \emph{Data \& Knowledge Engineering}, 2010.

\bibitem{kejriwal2013unsupervised}
M.~Kejriwal and D.~P. Miranker, ``An unsupervised algorithm for learning blocking schemes,'' in \emph{ICDM}.\hskip 1em plus 0.5em minus 0.4em\relax IEEE, 2013.

\bibitem{mugeni2023graph}
J.~B. Mugeni and T.~Amagasa, ``A graph-based blocking approach for entity matching using contrastively learned embeddings,'' \emph{SIGAPP}, 2023.

\bibitem{acluNoFly}
A.~Roe, ``Airline ``no fly'' lists trample the rights of people of color; seattle should not allow hotels to create a similar system,'' \url{www.aclu-wa.org}, 2020.

\bibitem{googleJobAdsBias}
A.~Peterson, ``Google’s algorithm shows prestigious job ads to men but not to women.'' \url{www.washingtonpost.com}, 2015.

\bibitem{hernandez1998real}
M.~A. Hern{\'a}ndez and S.~J. Stolfo, ``Real-world data is dirty: Data cleansing and the merge/purge problem,'' \emph{KDD}, 1998.

\bibitem{yang2020fairness}
K.~Yang, B.~Huang, J.~Stoyanovich, and S.~Schelter, ``Fairness-aware instrumentation of preprocessing pipelines for machine learning,'' in \emph{HILDA}, 2020.

\bibitem{papadakis2015schema}
G.~Papadakis, G.~Alexiou, G.~Papastefanatos, and G.~Koutrika, ``Schema-agnostic vs schema-based configurations for blocking methods on homogeneous data,'' \emph{PVLDB}, 2015.

\end{thebibliography}

\end{document}